\documentclass[11pt]{article}
\pdfoutput=1
\usepackage[margin=1in]{geometry}
\usepackage{graphicx}
\usepackage{subcaption} 
\usepackage{amsmath,amssymb,amsfonts}
\usepackage{booktabs}
\usepackage{algorithm}
\usepackage{algpseudocode}
\usepackage{textcomp}
\usepackage[numbers]{natbib}

\usepackage[expansion=false]{microtype}
\usepackage{multirow}

\usepackage[unicode, colorlinks=true, linkcolor=blue, citecolor=blue, urlcolor=blue, breaklinks=true]{hyperref}

\title{How Confident Is the First Token? An Uncertainty-Calibrated Prompt Optimization Framework for Large Language Model Classification and Understanding}

\author{
	Wei Chen$^{\dag}$ \quad Guoyang Ju$^{\dag}$ \quad Yuanyuan Qi$^{*}$ \\
	China Jiliang University, Hangzhou, China \\
	\texttt{\{wei.chen, guoyangju, qiyuanyuan\}@cjlu.edu.cn} \\
	\thanks{$^{\dag}$Equal contribution. $^{*}$Corresponding author.}
}

\date{}
\begin{document}
	\maketitle

\begin{abstract}
With the widespread adoption of large language models (LLMs) in natural language processing, prompt engineering and retrieval-augmented generation (RAG) have become mainstream paradigms for improving models' ability to handle complex tasks. However, LLMs typically generate outputs autoregressively, making uncertainty in the output inevitable. Since model performance is highly sensitive to prompt design, it is crucial to provide reliable evidence for prompt optimization via precise uncertainty measurement.

For multi-class multiple-choice (i.e., understanding) tasks, conventional uncertainty measures based on output probabilities (e.g., entropy) treat all classes equally and ignore differences in class priors in pretraining corpora. As a result, they cannot distinguish \emph{spurious confidence dominated by priors} from \emph{true certainty driven by contextual understanding}, leading to poor confidence calibration. To address this issue, we propose a first-token-based metric, \emph{Log-Scale Focal Uncertainty} (LSFU). Inspired by focal loss \citep{lin2017focal}, LSFU incorporates label prior probabilities as a risk-modulation factor, suppressing background noise from high-frequency, easy-to-classify classes while emphasizing prediction risk for long-tail, low-frequency classes; a dynamic weighting mechanism unifies the measurement scale.

Based on LSFU, we further propose an uncertainty-calibrated prompt optimization framework, \emph{UCPOF}. UCPOF leverages the first token of the model output: it uses LSFU to select high-quality exemplars and to dynamically optimize prompts. Comprehensive evaluations show that UCPOF improves average accuracy by 6.03\% over few-shot baselines and even surpasses always-on full RAG, boosting the overall average accuracy by 5.75\% while cutting the average retrieval trigger rate by 50.66\% . By adaptively triggering RAG only for high-uncertainty samples, our framework significantly lowers computational costs while maintaining state-of-the-art performance.

\end{abstract}

\section{Introduction}
With rapid advances in large language models (LLMs), classification and understanding tasks based on in-context learning (ICL) have become a mainstream paradigm in natural language processing~\citep{liu2023pre}. Under this paradigm, models can perform reasoning without parameter updates, relying only on prompts, thereby enabling genuine few-shot learning~\citep{perez2021true} and substantially reducing adaptation costs for downstream tasks. Recent studies further highlight the decisive role of the first predicted token in ICL inference. In LLM-based classification and multiple-choice tasks, the first predicted token occupies a pivotal position. Standard evaluation protocols (e.g., MMLU) predominantly rely on First-Token Probability (FTP) to derive model decisions, as it serves as the initial manifestation of the model's internal reasoning outcome. Recent studies also highlight the 'attention sink' phenomenon, where models concentrate significant representational mass on initial tokens to stabilize long-sequence generation. Zeng et al.~\citep{zeng2025pruning} show that the first token often carries core semantic information and can serve as a key signal for efficient inference. Snel and Oh~\citep{snel2025first} find that the first-token distribution differs markedly when distinguishing model ``hallucinations'' from ``true knowledge,'' offering an important entry point for reliability assessment. Unlike question answering, where answers may span multiple tokens, classification labels form a discrete, fixed set of categories, and the model's core decision depends on the first predicted token, which may directly correspond to the class label. Therefore, we posit the First-Token Confidence Hypothesis: the uncertainty manifested in the first token is a highly reliable, early-stage indicator of the model's overall task-understanding quality.
This makes first-token uncertainty more directly influential on classification outcomes and provides a precise lever for prompt optimization.

However, when pursuing higher performance, the model's raw predictive confidence, although positively correlated with accuracy overall, is still not sufficiently informative for long-tail distributions or borderline ambiguous cases. This leaves the prompt design without reliable quantitative evidence. Specifically, entropy computed from the output distribution has two major limitations and thus struggles to support effective prompt optimization. First, it ignores prior difficulty differences among classes. As noted by Kadavath et al.~\citep{kadavath2022language}, models often ``know what they know,'' yet in long-tail tasks they tend to be influenced by pretraining frequency~\citep{pinto2024fair}, assigning extremely high probability to high-frequency terms. Therefore, low entropy when predicting frequent terms and low entropy when predicting rare terminology do not indicate the same level of mastery; entropy-based exemplar selection may yield redundant high-frequency examples and miss representative low-frequency ones. Second, full-vocabulary entropy is computationally expensive and susceptible to tail noise. Although Clark et al.~\citep{clark2025well} show that first-token entropy can approximate token-level entropy, without calibrating prior bias this approximation remains insufficiently precise for complex classification tasks and cannot determine whether the model's certainty is driven by effective prompt guidance or by pretraining priors.

To improve prediction accuracy, retrieval-augmented generation (RAG) introduces external knowledge to provide semantic support for prompts~\citep{jiang2023active}. Nevertheless, existing RAG methods often adopt an ``always retrieve'' strategy, retrieving indiscriminately for all samples, which has two notable drawbacks and can even weaken prompt guidance. First, it increases the computational cost and inference latency, reducing the efficiency of prompt-iteration. Second, for easy samples that a high-quality prompt can handle on its own, retrieved but imperfectly matched context can become semantic noise, interfering with the model's understanding of the prompt's core intent and lowering accuracy~\citep{liu2024lost}. While Sharma et al.~\citep{sharma2025think} propose sequence-level entropy as a confidence signal, without leveraging class priors and first-token characteristics it is still difficult to precisely distinguish ``high-confidence samples that need no external knowledge'' from ``low-confidence samples that require external assistance.'' Therefore, the core challenge is to design a prompt-optimization-oriented metric that combines the efficiency of first-token semantic representations with class-prior distributions to overcome prior bias, support optimal exemplar selection for static prompts, and dynamically decide when prompt correction is needed, thus providing a quantitative criterion for knowledge augmentation and adaptive prompt construction.

We summarize our main contributions as follows:
\begin{enumerate}
  \item We propose the \emph{Log-Scale Focal Uncertainty} (LSFU) metric. Focusing on the distribution of the first output token, LSFU is inspired by focal loss and provides a prompt-optimization-oriented calibration measure that requires no additional training. LSFU incorporates the label prior distribution as a modulation factor to reweight conventional entropy, suppressing ``safe'' uncertainty signals from high-frequency classes while emphasizing prediction risk for low-frequency classes, thereby reducing evaluation noise caused by label bias and providing reliable quantitative evidence for exemplar selection and dynamic prompt correction.
  \item We propose an LSFU-driven static prompt optimization strategy (\emph{Gold Shot Selection}). Using LSFU, we select few-shot exemplars with the lowest LSFU values (i.e., those for which the model is most confident and whose predictions best align with prior intuition) from the sample set. Compared with random selection or similarity-based selection, this confidence-based strategy provides more stable classification anchors and significantly improves the in-context learning baseline, addressing the ``blindness'' of traditional prompt exemplar selection.
  \item We propose an uncertainty-aware dynamic prompt optimization framework, \emph{UCPOF} (Uncertainty-Perception Dynamic Prompt Optimization Framework). Centered on ``static-prompt inference + conditional dynamic prompt correction,'' UCPOF uses LSFU as an intelligent gate to trigger RAG retrieval only for high-risk samples. This mechanism substantially reduces inference cost while filtering noise introduced by always-on RAG, improving both efficiency and accuracy.
\end{enumerate}

\section{Related Work}
In this section, we review prior work on in-context learning and prompt optimization, uncertainty quantification for large language models, retrieval-augmented generation and adaptive inference, and we provide formal definitions relevant to our setting.

\subsection{In-Context Learning and Automated Prompt Optimization}
In-context learning (ICL) has become a central paradigm for deploying large language models (LLMs), thanks to its ability to adapt to downstream tasks \emph{without parameter updates} and \emph{solely through prompts}~\citep{shin2020autoprompt}. Automated prompt optimization is a key means of improving ICL performance, and its research landscape has been systematically summarized in recent surveys~\citep{ramnath2025systematic, xu2025systematic}. However, ICL performance is highly sensitive to prompt design across multiple dimensions, including exemplar selection~\citep{fu2024tise}, ordering~\citep{lu2022fantastically}, instruction formatting~\citep{liu2025beyond}, positional bias~\citep{cobbina2025show}, and semantic realization~\citep{wei2025integration}. Designing prompts that are both effective and stable therefore remains a major challenge.

Early prompt optimization research focused on \emph{soft prompts} in continuous embedding spaces. Such methods are parameter-efficient, and their scalability has been extensively validated~\citep{lester2021power}. For example, AutoPrompt~\citep{shin2020autoprompt} automatically generates discrete trigger tokens to elicit knowledge from pretrained models, while Prefix-Tuning~\citep{li2021prefix} prepends learnable continuous prefixes to the model input. Nevertheless, these prompts are often difficult to interpret, which limits their direct applicability in practical settings.

More recent work has increasingly shifted toward \emph{discrete prompt optimization}, exploring gradient-based search strategies~\citep{shin2020autoprompt2} as well as reinforcement learning and evolutionary algorithms~\citep{secheresse2025gaapo, hou2025improve, madaan2024self} to search for optimal prompt structures in discrete spaces. Liu et al.~\citep{liu2025beyond} propose jointly optimizing prompt content and instruction format, while Dong et al.~\citep{dong2025emergent} investigate intrinsic response-planning mechanisms to support dynamic prompt adjustment. Notably, the prototype framework proposed by Liu et al.~\citep{liu2024prompt} improves ICL stability by clustering in feature space and selecting representative samples as prompt exemplars, which aligns with our goal of building prompts by selecting highly typical examples.

Despite these advances, existing automated prompt optimization methods have a key limitation: most treat accuracy as the sole objective and ignore confidence calibration during inference~\citep{liu2024prompt}. This lack of calibration can lead to \emph{overconfidence} that does not reflect true capability in long-tail or complex-sample scenarios, leaving exemplar selection without quantitative guidance and dynamic adjustment without reliable signals. Our proposed LSFU metric and prompt optimization strategies aim to fill this gap by providing calibration-guided prompt optimization.

\subsection{Uncertainty Quantification for LLMs}
Uncertainty quantification is a core tool for assessing model reliability, and it is also a key basis for deciding \emph{whether prompt optimization is needed}. Accurate uncertainty measures provide quantitative criteria for prompt exemplar selection and dynamic correction. Existing methods can be broadly categorized into two groups.

\textbf{(1) Consistency-based methods.} These methods estimate confidence by sampling multiple reasoning paths or introducing self-feedback mechanisms. For instance, self-consistency~\citep{wang2023self} measures agreement across multiple sampled paths, and Madaan et al.~\citep{madaan2024self} use the model to score its own generations. While such approaches can improve confidence estimation, their inference cost grows linearly with the number of samples, making them unsuitable for real-time prompt optimization and high-throughput settings.

\textbf{(2) Logit-based methods.} These methods compute uncertainty directly from the model's output probability distribution, offering high computational efficiency and thus being a natural fit for prompt optimization. Lin et al.~\citep{lin2023generating} study uncertainty quantification under black-box access, addressing scenarios without internal model visibility. Since full-vocabulary entropy is expensive and noisy, entropy over top-$K$ logits is a common approximation~\citep{clark2025well}. Kuhn et al.~\citep{kuhn2023semantic} observe that conventional entropy can be confounded by synonym competition and propose semantic entropy to improve evaluation. Further studies by Clark et al.~\citep{clark2025well} and Cappelletti et al.~\citep{cappelletti2025improving} suggest that focusing on the \emph{first-token} distribution can improve the precision of uncertainty estimation while preserving efficiency, aligning with our first-token-centric design.

Nevertheless, most logit-based metrics ignore the label prior distribution~\citep{sharma2025think}. In long-tail tasks, such prior bias prevents existing metrics from distinguishing ``reasonable hesitation on low-frequency hard cases'' from ``overconfidence on high-frequency easy cases,'' and thus from providing reliable guidance for prompt optimization. Our proposed Log-Scale Focal Uncertainty (LSFU) explicitly incorporates label priors to calibrate confidence and address this limitation.

\subsection{Retrieval-Augmented Generation and Dynamic Inference}
Retrieval-augmented generation (RAG) enriches prompts with external knowledge bases to mitigate hallucinations and knowledge limitations in LLMs~\citep{jiang2023active}, making it an important pathway for dynamic prompt optimization. However, existing RAG approaches face clear bottlenecks. Always-on RAG retrieves external knowledge indiscriminately for all samples, which not only incurs substantial computational cost and latency~\citep{liu2024lost} but can also amplify semantic noise due to LLMs' long-context degradation~\citep{lu2024insights}. For simple samples, such retrieval can introduce redundant distractions into prompts, undermining the model's understanding of the prompt's core intent and reducing accuracy~\citep{dong2025emergent}.

To balance efficiency and performance, \emph{adaptive} RAG has emerged, with the key idea of triggering retrieval \emph{only} for high-uncertainty samples to enable targeted prompt correction. Active RAG~\citep{jiang2023active} selects samples for retrieval using active learning strategies. Yuan et al.~\citep{yuan2024instance} use zero-shot chain-of-thought to adapt prompts instance-wise. Dong et al.~\citep{dong2025emergent} explore intrinsic response planning to decide when retrieval is needed. However, a major limitation of many dynamic frameworks is inaccurate gating: triggering signals often fail to account for label-prior bias, making it hard to determine whether external knowledge is necessary. This can lead to ineffective retrieval (unnecessary corrections for easy samples) or missed retrieval (insufficient optimization for difficult samples). Methods such as LLM2LLM~\citep{lee2024llm2llm} further enhance correction via iterative interactions, but introduce additional overhead and are difficult to deploy in real-time settings.

In contrast, our uncertainty-aware dynamic prompt optimization framework, UCPOF, uses a lightweight LSFU metric as an intelligent gate to trigger retrieval and prompt correction only when necessary. By combining ``static prompt optimization + dynamic prompt correction,'' UCPOF improves both prompt effectiveness and inference efficiency, making it more suitable for practical applications.

\subsection{Formal Definitions}
We now provide formal definitions for in-context learning (ICL), retrieval-augmented generation (RAG), and the decision rule used in our framework.

\textbf{In-context learning (ICL).} Given a pretrained language model $P_{\theta}$, a test input $x$, and a set of demonstration examples $C=\{(x_1,y_1),\ldots,(x_k,y_k)\}$, ICL predicts the label $y$ by maximizing its conditional probability:
\begin{equation}
P_{\mathrm{ICL}}(y\mid x,C)=\prod_{t=1}^{T} P_{\theta}(y_t \mid C,x,y_{<t}).
\end{equation}
Here, the label $y$ is a token sequence $(y_1,\ldots,y_T)$. Prior studies by Olsson et al.~\citep{olsson2022context} and Zeng et al.~\citep{zeng2025pruning} show that the first generated token often contains decisive information for subsequent decisions (e.g., via \emph{induction heads}) and can already support the core judgment in classification tasks. This property motivates first-token-based uncertainty measures (such as our LSFU), which can support prompt optimization efficiently without redundantly computing distributions over the full output sequence.

\textbf{Retrieval-augmented generation (RAG).} Given an external knowledge base $K$, a retriever $R$ retrieves relevant documents according to the input $x$, i.e., $z = R(x,K)$. The generation probability under RAG is:
\begin{equation}
P_{\mathrm{RAG}}(y\mid x,z)=P_{\theta}(y\mid x,z).
\end{equation}
The key idea is to improve the semantic guidance of the prompt by augmenting it with external knowledge $z$, while carefully controlling when retrieval is triggered to avoid introducing noise.

\textbf{UCPOF decision rule.} The proposed UCPOF framework dynamically selects the prompt optimization path based on the score $S(x)$:
\begin{equation}
P_{\mathrm{UCPOF}}(y\mid x)=
\begin{cases}
P_{\mathrm{ICL}}(y\mid x,C^{\ast}) & \text{if } S(x) < \tau,\\
P_{\mathrm{RAG}}(y\mid x,C^{\ast},z) & \text{if } S(x) \ge \tau.
\end{cases}
\end{equation}
Here, $S(x)$ is the Log-Scale Focal Uncertainty (LSFU) score, $\tau$ is a dynamic threshold, and $C^{\ast}$ is a statically optimized prompt constructed from gold-shot exemplars selected using LSFU.

\section{Theoretical Motivation and Analysis}
\subsection{First-Token Confidence Hypothesis}
In an autoregressive language model, given an input $x$, the generation probability of a target sequence $y=(y_1,\ldots,y_T)$ can be factorized as
\begin{equation}
P(y \mid x)=\prod_{t=1}^{T} P(y_t \mid x, y_{<t}),
\end{equation}
where the conditional distribution of the first token, $P(y_1\mid x)$, sets the initial semantic direction of the entire generation trajectory.

From an information-theoretic perspective, if $P(y_1\mid x)$ is highly concentrated (i.e., low entropy and a prominent peak), the model has formed a clear decision within the candidate semantic space; conversely, a diffuse distribution indicates substantial uncertainty among multiple competing semantic paths.

Based on this, we propose the \emph{First-Token Confidence Hypothesis}: the concentration of the first-token predictive distribution can serve as an early proxy for the model's overall predictive confidence. This hypothesis is motivated by the cascading nature of autoregressive generation: uncertainty at the initial decision stage is often amplified during subsequent generation, thereby increasing overall prediction risk. Therefore, first-token confidence provides an efficient and theoretically grounded early risk signal.

In this work, we construct an uncertainty measure using only the information in the first-token distribution, enabling early confidence assessment and guiding subsequent prompt optimization and the selection of inference strategies.

\subsection{Properties of the Proposed Uncertainty Metric}
Let the first-token predictive distribution be $p = P(y_1 \mid x)$. The uncertainty function $U(x)$ defined in this paper satisfies the following key properties:
\begin{enumerate}
  \item \textbf{Monotonicity.} When $p$ becomes more concentrated (e.g., $\max_i p_i$ increases and entropy decreases), $U(x)$ decreases monotonically, remaining consistent with increasing model confidence.
  \item \textbf{Boundedness.} Since the vocabulary is finite, $U(x)$ has theoretical upper and lower bounds, ensuring comparability across models and tasks and supporting threshold-based decisions.
  \item \textbf{Stability.} $U(x)$ responds continuously and in a bounded manner to small perturbations of the probability distribution, making it robust to estimation noise and preventing strategy oscillations caused by local fluctuations.
  \item \textbf{Connection to classical metrics.} Under certain forms, $U(x)$ can reduce to traditional uncertainty measures based on entropy or confidence margins. By explicitly modeling distributional structure, our method can more sensitively capture semantic disagreement at the first-token stage and is better suited for early uncertainty assessment.
\end{enumerate}
Together, these properties ensure that the proposed metric is both theoretically interpretable and stable for practical deployment.

\subsection{Risk-Aware Prompt Decision Interpretation}
We formulate prompt optimization as a risk-aware decision problem: the system first estimates the prediction risk for a given sample based on $U(x)$, and then adaptively chooses an inference strategy. Specifically, when $U(x)$ is below a threshold, the model is regarded as highly reliable and a basic prompt is used directly to save computation; when $U(x)$ exceeds the threshold, an enhanced strategy (e.g., prompt reconstruction, retrieval augmentation, or multi-step verification) is triggered to suppress potential errors.

This mechanism can be viewed as a practical approximation of \emph{selective prediction}, dynamically trading off prediction risk against computational cost. Selective classification, as its core theoretical foundation, has been well validated in deep neural networks~\citep{geifman2017selective}. If $U(x)$ can effectively distinguish ``easy'' samples from ``hard'' samples, then allocating additional resources only to high-risk samples can significantly reduce the average inference cost while maintaining performance. Therefore, the proposed first-token-uncertainty-driven mechanism not only provides an interpretable, confidence-guided pathway for prompt optimization, but also lays a theoretical foundation for building efficient and adaptive LLM inference systems.

\section{Method}
\subsection{Log-Scale Focal Uncertainty (LSFU)}
In LLM classification tasks, conventional entropy-based uncertainty measures often fail to provide reliable confidence calibration. In particular, models tend to output high-frequency tokens from the pretraining corpus (high prior); such high-frequency outputs are frequently accompanied by very low entropy, yet can still manifest as ``overconfidence.'' This issue leads to two core difficulties for prompt design: (1) static exemplar selection lacks quantitative evidence and may mistakenly select redundant high-frequency samples or borderline ambiguous samples; (2) dynamic adjustment lacks reliable signals, making it difficult to precisely decide when external knowledge should be introduced to optimize the prompt.

To address these issues, we shift uncertainty measurement from a notion of ``information disorder'' to \emph{prompt-optimization-oriented prediction risk assessment}, and propose the \emph{Log-Scale Focal Uncertainty} (LSFU) metric. LSFU combines the model's internal semantic confusion with the external label-prior distribution, aiming to provide a unified and precise quantitative criterion for both static exemplar screening and dynamic prompt-correction triggering. We define LSFU as
\begin{equation}
\mathrm{LSFU}(x,y)=\log_{10}\!\Big(H(P_{\mathrm{top}\text{-}k})\cdot(1-P_{\mathrm{prior}}(y))^{2}+\varepsilon\Big),
\end{equation}
where $\varepsilon$ is a small constant (set to $10^{-8}$ in this work) for numerical stability.

\textbf{Theoretical Insights.} The formulation in Eq. (5) directly instantiates the theoretical principles of monotonicity and boundedness established in Section 3. By modulating the uncertainty with label priors, LSFU effectively suppresses the noise from high-frequency labels in the pre-training corpus, ensuring that the uncertainty value faithfully reflects the model's task-specific comprehension. This provides a rigorous mathematical foundation for the threshold-based gating in our framework.

We next explain the components and their relevance to prompt optimization. Let $x$ denote the input sequence and $y$ the predicted class. Let $H(P_{\mathrm{top}\text{-}k})$ be the Shannon entropy computed over the top-$K$ candidate tokens in the first-token distribution (derived from the final-layer logits). In our experiments, we set $K=50$; empirical validation (Appendix~C) shows that the entropy values change negligibly when using $K\in\{50,100,500\}$ or the full vocabulary, and $K=50$ offers a good trade-off between coverage and efficiency for real-time prompt optimization. This entropy directly reflects the model's ``hesitation,'' and forms the basis for deciding whether a sample is suitable as a prompt exemplar and whether prompt correction is needed.

To ensure completeness of the probability space and avoid distorted optimization signals, probabilities within the top-$K$ set are re-normalized. Let $p_i$ be the original probability of the $i$-th candidate token; we define the re-normalized probability as
\begin{equation}
\hat{p}_i=\frac{p_i}{\sum_{j\in V_K} p_j}, \quad i\in V_K.
\end{equation}
Based on this, the entropy $H(P_{\mathrm{top}\text{-}k})$ is computed as
\begin{equation}
H(P_{\mathrm{top}\text{-}k})=-\sum_{i\in V_K} \hat{p}_i\log_2 \hat{p}_i.
\end{equation}

The term $P_{\mathrm{prior}}(y)$ is the prior probability of the predicted class $y$ under the training distribution, representing its real-world frequency. Its key role is to calibrate class weights for prompt optimization: it prevents high-frequency classes from being over-selected as prompt exemplars due to pretraining inertia, while ensuring that representative cases from low-frequency classes are not missed, addressing the common ``favor frequent classes, neglect long-tail'' bias in exemplar selection.

The factor $(1-P_{\mathrm{prior}}(y))^{2}$ is a risk-modulation term inspired by focal loss~\citep{lin2017focal}. For high-prior classes ($P_{\mathrm{prior}}(y)\to 1$), it approaches $0$, filtering out small entropy fluctuations caused by synonym competition and preventing redundant high-frequency samples from being mistakenly selected as exemplars; it also helps avoid unnecessarily triggering dynamic correction for easy samples that can already be reliably handled by a static prompt. For low-prior classes ($P_{\mathrm{prior}}(y)\to 0$), it approaches $1$, highlighting prediction risk on long-tail samples so that hard cases can be identified: they are more likely to be included as typical exemplars in static prompts, and can also trigger RAG-based knowledge augmentation and reflective prompt construction when needed.

The $\log_{10}$ transform avoids numerical compression and preserves discriminability between low- and high-risk cases. The $\varepsilon$ term is a small constant (set to $10^{-8}$ in this work) to prevent numerical overflow when $H(P_{\mathrm{top}\text{-}k})\cdot(1-P_{\mathrm{prior}}(y))^{2}$ approaches $0$.

\subsection{Static Optimization: LSFU-based Gold Shot Selection}
Constructing high-quality few-shot prompts is a prerequisite for improving the performance of in-context learning (ICL)~\citep{liu2025beyond, liu2024prompt}. Traditional prompt-exemplar selection typically relies on random sampling or semantic-similarity matching, without a quantitative evaluation of an exemplar's ``guidance value.'' This can introduce borderline ambiguous or semantically ambiguous samples, increasing the model's cognitive load and blurring the decision boundary. To address this issue, we propose an LSFU-based static prompt optimization strategy: we use LSFU to select \emph{gold-shot} exemplars and construct a static prompt with strong guidance and low noise, laying the foundation for subsequent dynamic optimization.

The key criterion for gold-shot selection is to choose $N$ samples from the dataset that are \emph{predicted correctly} and have the \emph{lowest LSFU} values. The detailed procedure is shown in Algorithm~\ref{alg:gold-shot}. A low LSFU indicates that a sample simultaneously satisfies two desirable properties---high intrinsic model confidence and good alignment with the prior distribution. As prompt exemplars, such samples offer three advantages that directly support prompt optimization:
\begin{enumerate}
  \item \textbf{Clear guidance boundaries.} Samples with the lowest LSFU are typically either low-risk high-frequency classes or representative low-frequency classes; they tend to lie near the cluster centers of their classes in feature space. Embedding such samples as exemplars provides clear decision anchors, helping the model quickly capture the core semantics of the task and avoiding blurred guidance caused by boundary cases.
  \item \textbf{Minimized cognitive noise.} Compared with hard boundary samples, high-confidence samples are semantically clearer and less ambiguous. Including them in the prompt significantly reduces the model's cognitive burden and prevents ambiguous exemplars from misleading the model when it is reasoning about easy test samples, thereby preserving the prompt's core intent.
  \item \textbf{Purified supervision signals.} Low LSFU corresponds to a sharp predictive distribution, indicating low semantic ambiguity and clear class membership. As ``teaching examples'' in the prompt, such samples provide the cleanest supervision signals and improve the reliability of prompt guidance~\citep{liu2024prompt}.
\end{enumerate}

\begin{algorithm}[t]
\caption{LSFU-based Gold-Shot Exemplar Selection (Static Prompt Construction)}
\label{alg:gold-shot}
\begin{algorithmic}[1]
\Require Dataset $D=\{(x_1,y_1),\ldots,(x_M,y_M)\}$; pretrained LM $P_{\theta}$; number of gold shots $N$ (typically 4--8); seed size $K$ (set to $K=N$); instruction template $\mathcal{T}$ (Example$\rightarrow$Task$\rightarrow$Input); top-$K$ entropy parameter $K_{\mathrm{top}}=50$
\Ensure Statically optimized prompt $P_{\mathrm{static}}$ containing $N$ gold-shot exemplars
\State Randomly sample $K$ training instances $D_{\mathrm{seed}}\subseteq D$ and form $C_{\mathrm{seed}}=\{(x_i,y_i)\mid (x_i,y_i)\in D_{\mathrm{seed}}\}$
\State Construct initial prompt $P_{\mathrm{initial}} \gets \mathcal{T}(C_{\mathrm{seed}})$
\State Initialize $\texttt{valid\_samples}\gets[\ ]$
\ForAll{$(x,y)\in D\setminus D_{\mathrm{seed}}$}
  \State Run $P_{\theta}$ with prompt $P_{\mathrm{initial}}$ to obtain predicted label $\hat{y}$ and first-token logits distribution $L$
  \If{$\hat{y}=y$}
    \State Compute normalized top-$K$ entropy $H(P_{\mathrm{top}\text{-}k})$ from the top-$K_{\mathrm{top}}$ candidates in $L$
    \State Compute prior probability $P_{\mathrm{prior}}(y)=\#\{(x',y')\in D: y'=y\}/M$
    \State Compute $\mathrm{LSFU}(x,y)$ using Eq.~(5)
    \State Append $(x,y,\mathrm{LSFU}(x,y))$ to \texttt{valid\_samples}
  \EndIf
\EndFor
\State Sort \texttt{valid\_samples} in ascending order of LSFU
\State Select the top $N$ samples to form $C_{\mathrm{gold}}=\{(x_1,y_1),\ldots,(x_N,y_N)\}$
\State Construct $P_{\mathrm{static}} \gets \mathcal{T}(C_{\mathrm{gold}})$
\State \Return $P_{\mathrm{static}}$
\end{algorithmic}
\end{algorithm}

\subsection{Uncertainty-Calibrated Prompt Optimization Framework (UCPOF)}
Building on the core advantages of LSFU, we construct an \emph{uncertainty-calibrated prompt optimization framework} (UCPOF). UCPOF targets prompt optimization throughout the inference lifecycle. Inspired by the progressive human cognition pattern of ``quick judgment, then precise correction,'' UCPOF treats ``efficient inference with a statically optimized prompt'' as a mandatory base stage (fast thinking), providing stable guidance anchors for all samples. It then treats ``precise correction with a reflective prompt'' as a conditional refinement stage (slow thinking), optimizing only samples that are insufficiently guided by the static prompt. LSFU serves as an intelligent gate for deciding whether refinement should be activated, ensuring ``baseline guidance is always present, and dynamic correction is never blind.''

\begin{figure}[t]
  \centering
  \includegraphics[width=0.92\linewidth]{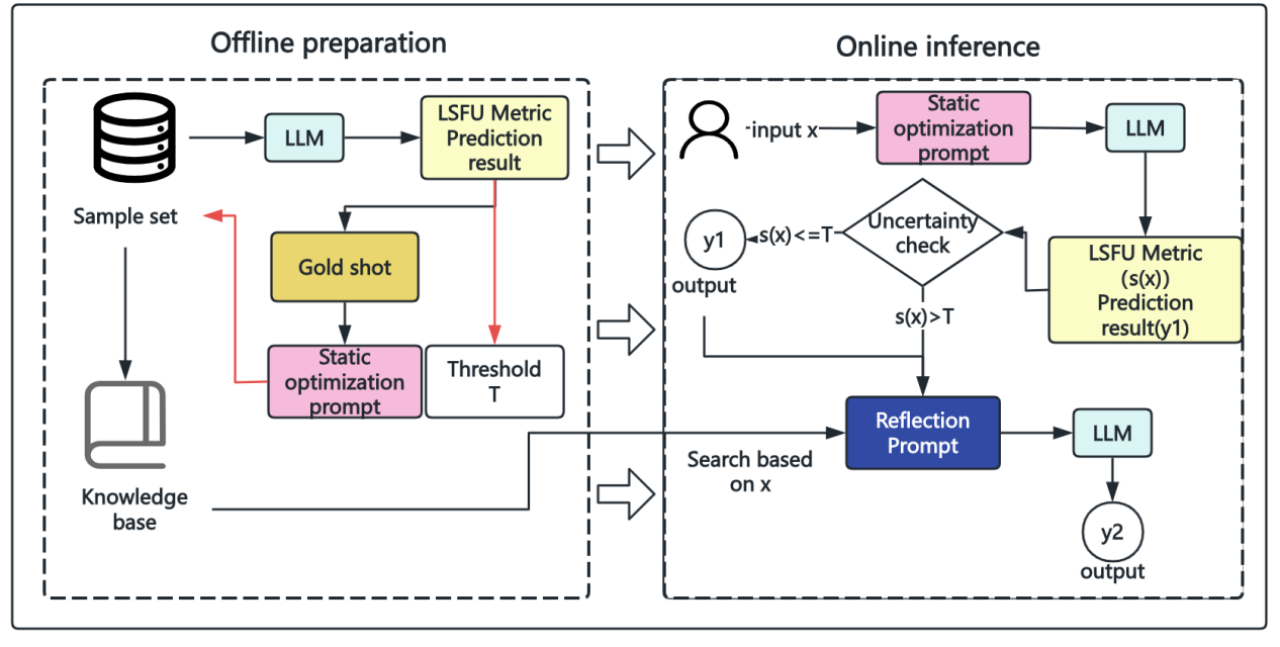}
  \caption{Schematic overview of the UCPOF workflow (offline preparation and online inference).}
  \label{fig:ucpof-overview}
\end{figure}

The overall UCPOF workflow consists of two stages: \emph{offline preparation} and \emph{online inference}. Both stages are organized around prompt optimization; a schematic overview is shown in Fig.~\ref{fig:ucpof-overview}.

\subsubsection{Offline Preparation}
The core goal of the offline stage is to provide, for the online inference stage, (i) a high-quality statically optimized prompt and (ii) a dedicated gating threshold $T$ tailored to that prompt. These two components are treated as a single decision unit for prompt optimization. Meanwhile, a vector knowledge base is constructed from all samples to support subsequent dynamic optimization. The steps are as follows:
\begin{enumerate}
  \item \textbf{Gold-shot selection and static prompt construction.} Based on the LSFU-driven screening strategy (Sec.~4.2), we select $N$ ``gold-shot'' exemplars from the training set with the lowest LSFU values and the highest predictive certainty, and construct a statically optimized prompt. This prompt has strong guidance and low noise, enabling efficient inference for easy samples without additional retrieval. The detailed procedure is provided in Sec.~4.2 and Algorithm~\ref{alg:gold-shot}.
  \item \textbf{Setting the gating threshold $T$.} The threshold $T$ directly determines the precision of dynamic prompt adjustment. A loose threshold may trigger retrieval for many low-risk samples (introducing noise), while an overly strict threshold may miss high-risk samples (leaving them uncorrected). We determine $T$ in two steps. First, we run inference with the statically optimized prompt over all samples in the dataset (excluding the gold-shot exemplars), and record for each sample the prediction correctness (correct/incorrect) together with its LSFU value $S(x)=\mathrm{LSFU}(x,y_i)$ (where $y_i$ is the predicted label under the static prompt). Second, we focus on the LSFU distribution of incorrectly predicted samples and choose a threshold that covers at least 90\% of these errors (as described in Sec.~5.2), and set this score as the gating threshold $T$. Under the assumption that uncertainty distributions are relatively stable within the same domain~\citep{jiang2023active}, the threshold optimized on the dataset can be transferred to unseen test sets of the same type. Even under mild distribution shift, a 90\% error-coverage target provides a sufficient safety margin, improving the robustness of the prompt correction mechanism.
  \item \textbf{Vector knowledge base construction.} Starting from the raw dataset, we parse each instance into its input sentence and ground-truth label, filter abnormal samples with invalid labels or missing sentences, and keep valid training samples as the core knowledge-base data. We use the lightweight pretrained encoder \texttt{all-MiniLM-L6-v2} to batch-encode sentences, converting each sentence into a 384-dimensional dense vector and applying vector normalization to improve the precision of subsequent similarity matching. We then build a vector knowledge base that contains the original sentences, the corresponding labels, and the normalized vectors.
\end{enumerate}

\subsubsection{Online Inference}
\begin{figure}[t]
  \centering
  \includegraphics[width=0.92\linewidth]{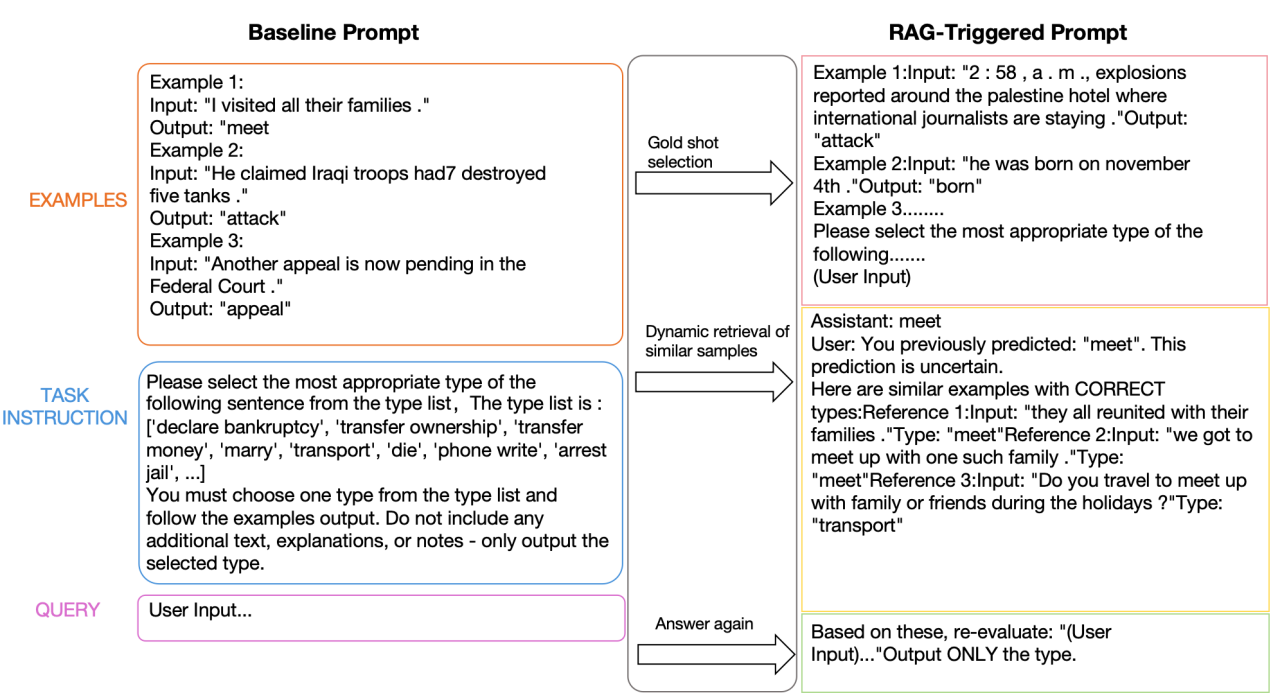}
  \caption{Structure of the reflection prompt for conditional correction.}
  \label{fig:reflection_prompt}
\end{figure}

The core of the online stage is to use the statically optimized prompt as the baseline and the threshold $T$ as the criterion: low-risk samples are handled directly with the static prompt for efficient inference, while high-risk samples trigger reflective-prompt construction and correction. This stage is the core execution flow of UCPOF and consists of three progressive steps:
\begin{enumerate}
  \item \textbf{Step 1: Static-prompt-guided inference (mandatory path; fast thinking).} For a test sample $x$, we first place it into the statically optimized prompt constructed in the offline stage to guide the LLM for a single prediction, producing an initial label $y_1$ (the class label in a classification task). We simultaneously compute the LSFU score $S(x)=\mathrm{LSFU}(x,y_1)$ to quantify the ``guidance reliability'' of the current static prompt on sample $x$---this is the key basis for deciding whether further prompt optimization is needed.
  \item \textbf{Step 2: Adaptive gating decision.} We compare $S(x)$ with the threshold $T$ to decide whether to dynamically optimize the prompt. If $S(x)\le T$, the static prompt is considered sufficient to guide a reliable prediction; we output $y_1$ directly as the final result to maximize inference efficiency. If $S(x)>T$, the static prompt is insufficient for this sample (e.g., due to long-tail difficulty or semantic ambiguity); we proceed to Step~3, retrieving external knowledge to construct a reflective prompt and correct the prediction bias.
  \item \textbf{Step 3: Reflective-prompt construction and corrected inference (conditional path; slow thinking).} When the gate is triggered, the system retrieves the top-$K$ most semantically similar reference samples and their ground-truth labels from the knowledge base given the input $x$. Using a structured template (Fig.~\ref{fig:reflection_prompt}) with ``initial-prediction notice + high-quality reference exemplars + explicit correction instruction,'' we embed the retrieved references into the prompt to form a reflective prompt. Inspired by reflective language agents~\citep{shinn2024reflexion}, this design first informs the model that the initial prediction may be uncertain, then supplements semantic guidance with high-quality in-class exemplars, and finally provides explicit instructions to re-judge based on the new information, mitigating semantic confusion. We then combine the reflective prompt with $x$ and run a second inference to obtain the final corrected prediction $y_2$.
\end{enumerate}

\section{Theoretical Analysis and Validation}
\subsection{Effectiveness of the LSFU Metric}
To better match the characteristics of classification tasks, LSFU focuses on the top-$K$ entropy of the first token. On the one hand, the number of categories in classification tasks is much smaller than the answer space of QA tasks, and an appropriate $K$ is sufficient to cover all possible categories and high-frequency interference terms. On the other hand, the labels of classification tasks are clear category identifiers, and the prior probability ($P_{\mathrm{prior}}$) can be directly calculated from the sample set without dealing with the semantic ambiguity of answers as in QA tasks, making the role of the risk modulation factor $(1-P_{\mathrm{prior}})^2$ more direct and controllable.

Inspired by the core idea of focal loss~\citep{lin2017focal}, LSFU reshapes the model's ``semantic focus'' by introducing the prior modulation factor $(1-P_{\mathrm{prior}})^2$. This mechanism is mathematically manifested as nonlinear risk weighting for samples of different frequencies, which is specifically reflected in the following two dimensions:
\begin{enumerate}
  \item \textbf{Prior Bias Suppression}: Even when entropy is slightly higher, if the model predicts a high-frequency category (high prior), the modulation factor decays rapidly due to the square term. This is equivalent to building a filter, which effectively suppresses the ``pseudo-uncertainty'' generated by the model due to pretraining statistical inertia (such as high-frequency synonym competition), and avoids false positives for low-risk samples.
  \item \textbf{Long-tail Risk Amplification}: For low-frequency categories (low prior), the modulation factor approaches 1, forcing the model to have extremely high certainty when processing long-tail samples. From a Bayesian perspective, the occurrence of rare events requires strong evidence support. LSFU amplifies tiny logical swings here, enabling the model to accurately capture those error risks that ``although the entropy value seems normal, there is insufficient evidence to support'', thus realizing early warning of long-tail hard cases.
\end{enumerate}

Notably, even in an experimental environment with a relatively balanced category distribution, introducing the prior modulation factor remains statistically meaningful. It can effectively offset the pretraining bias of the model, ensuring that the LSFU metric focuses on the instantaneous semantic features excited by the current prompt, rather than residual statistical inertia. In addition, through the nonlinear mapping of the square term, the metric stretches the dynamic range of the uncertainty signal at the decision boundary, which is highly consistent with the core requirements of uncertainty calibration and selective generation \citep{zablotskaia2023uncertainty}. This enables the system to maintain higher alertness when processing semantically ambiguous samples, thus providing a more sensitive trigger threshold for scheduling two-stage reasoning.

\begin{figure}[t]
  \centering
  \includegraphics[width=0.92\linewidth]{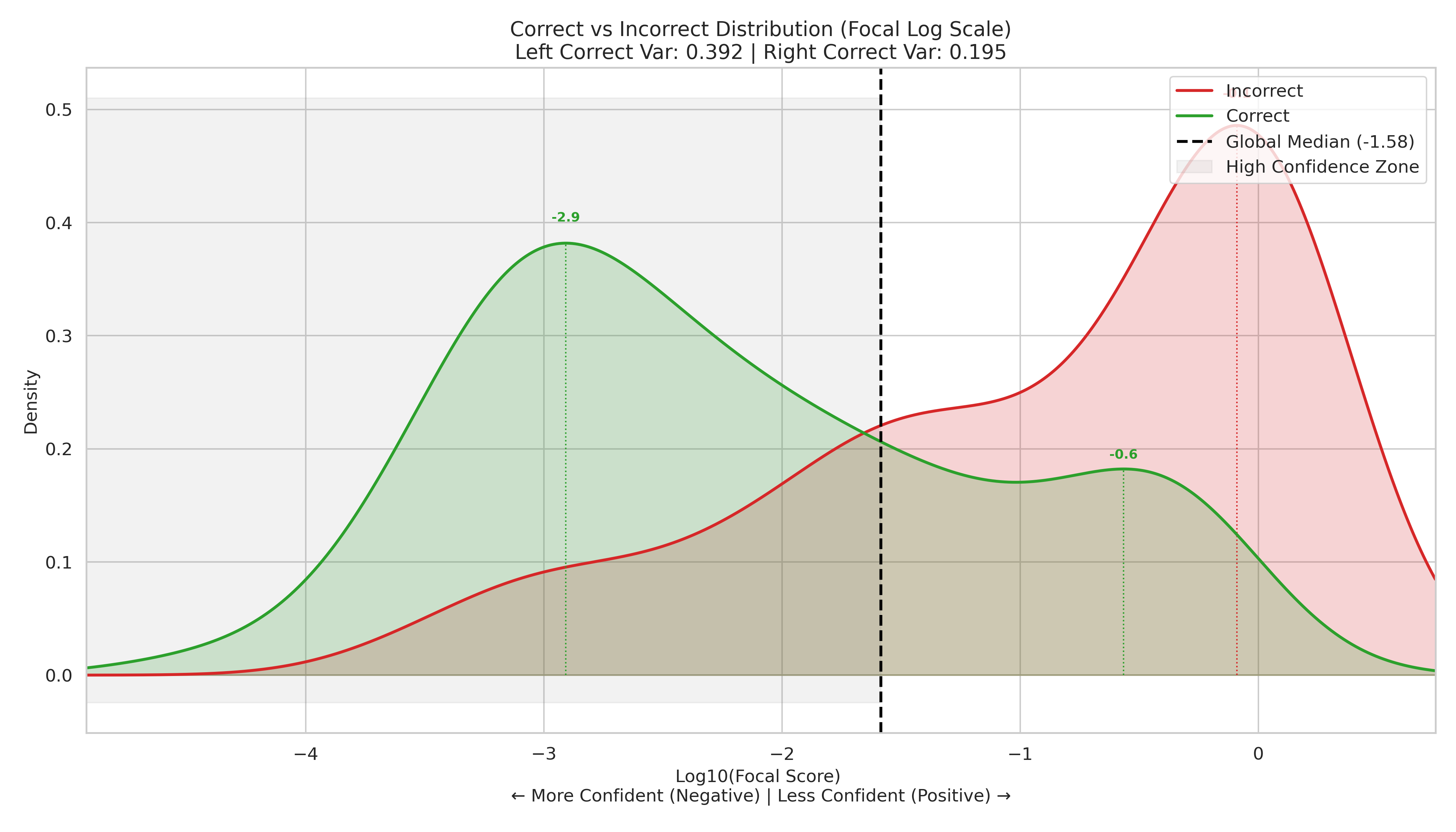}
  \caption{Distribution density of LSFU values for correct and incorrect predictions.}
  \label{fig:distribution_density}
\end{figure}

To intuitively verify LSFU's ability to characterize the model's prediction state, we correlate each sample's LSFU value with prediction correctness and plot the corresponding density distributions (Fig.~\ref{fig:distribution_density}). Both distributions lie in the negative LSFU range, yet exhibit clear separation along the horizontal axis. The distribution center of correct samples (green curve) is shifted leftward (corresponding to higher model confidence), while that of incorrect samples (red curve) is shifted rightward (corresponding to lower confidence). Although there is a certain overlap, this separation trend confirms LSFU's effectiveness in characterizing the model's prediction state.

\begin{figure}[t]
  \centering
  \begin{subfigure}{0.48\textwidth}
    \centering
    \includegraphics[width=\linewidth]{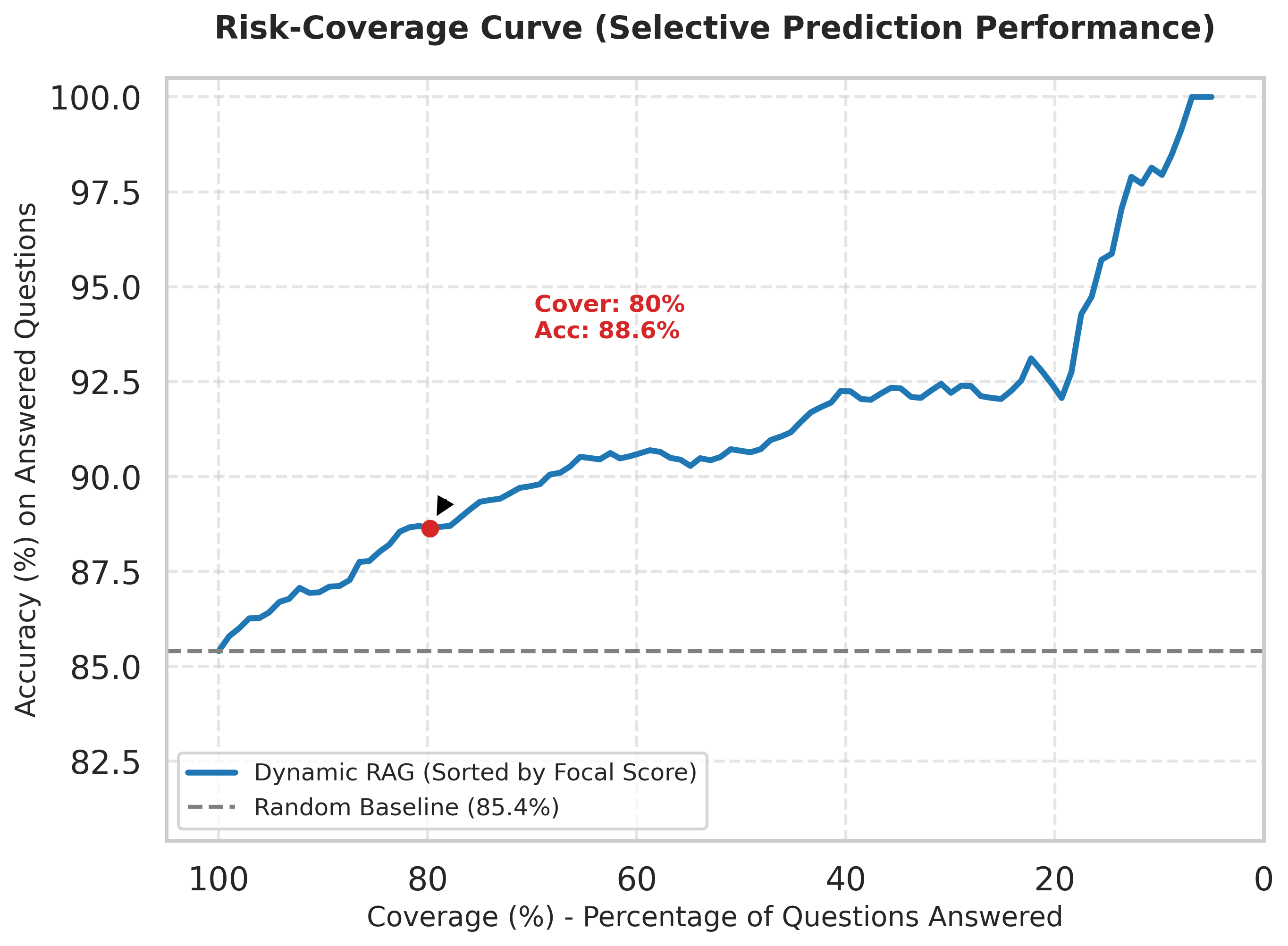}
    \caption{CASIE Dataset}
    \label{fig:risk_coverage_casie}
  \end{subfigure}
  \hfill 
  \begin{subfigure}{0.48\textwidth}
    \centering
    \includegraphics[width=\linewidth]{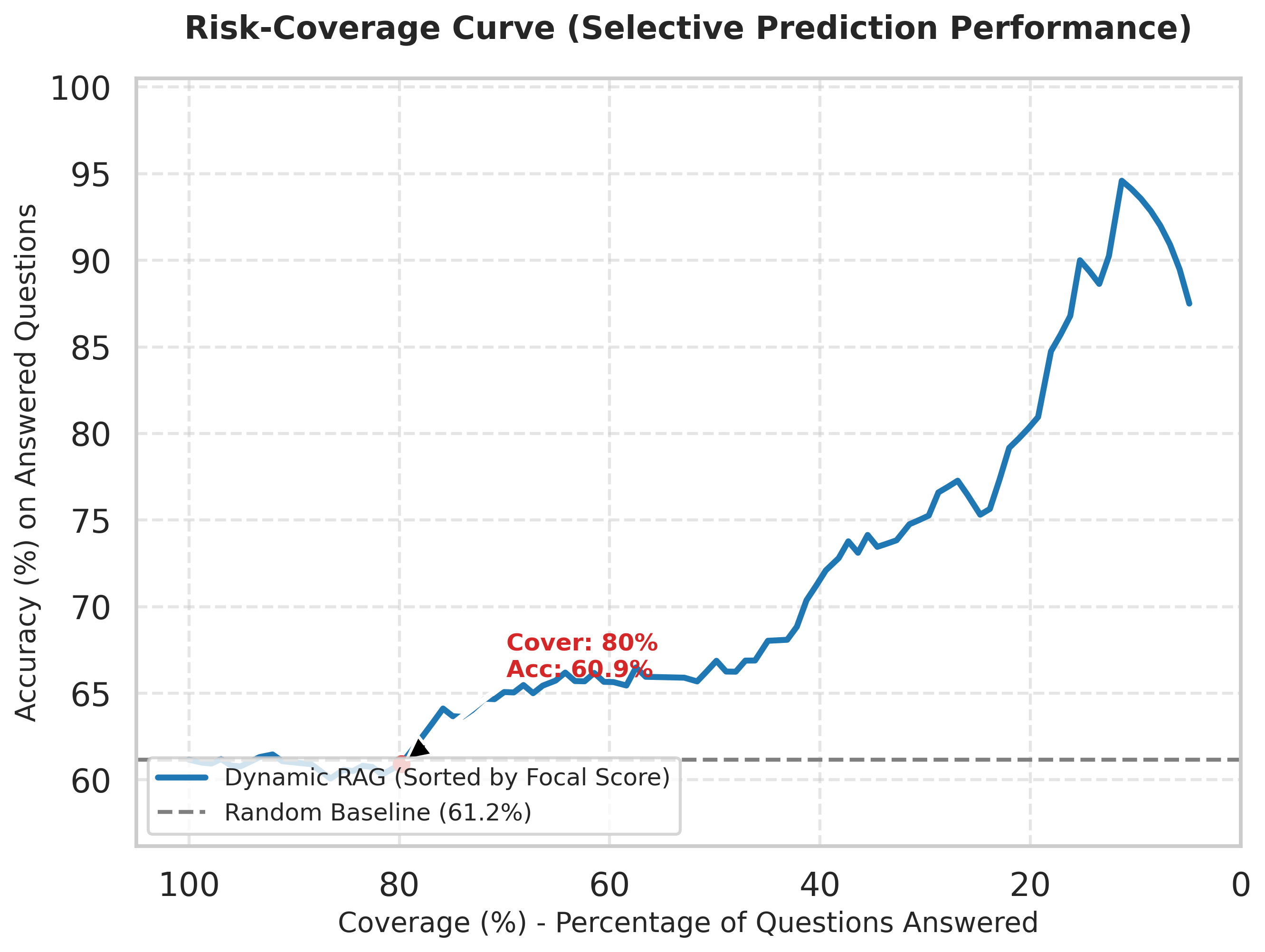}
    \caption{ACE Dataset}
    \label{fig:risk_coverage_ace}
  \end{subfigure}
  \caption{Risk-coverage curves of LSFU-based selective prediction on CASIE and ACE datasets.}
  \label{fig:risk_coverage}
\end{figure}

To further verify the core role of LSFU in selective prediction, we plot the risk-coverage curve (Fig.~\ref{fig:risk_coverage}), which shows the evolution trend of the accuracy of the remaining samples when the system refuses to answer according to the uncertainty score (i.e., the coverage rate decreases).

\textbf{Analysis and findings.}
\begin{enumerate}
  \item \textbf{Monotonically increasing accuracy.} As shown in Fig.~\ref{fig:risk_coverage}, as the coverage rate decreases (i.e., samples with high LSFU scores are eliminated), accuracy shows a significant upward trend. This indicates that Log-Scale Focal Uncertainty can effectively rank sample difficulty: higher scores correspond to higher error risk, which is a prerequisite for the dynamic framework.
  \item \textbf{Clear safety boundary.} On the CASIE dataset, when coverage drops to 10\%, accuracy reaches 88.6\%, suggesting that the metric identifies a reliable safety region. On the highly challenging ACE dataset, although the baseline accuracy is low ($\sim$51\%), accuracy increases to 60.9\% after eliminating 20\% of high-risk samples. These results indicate that LSFU can identify the ``safe coverage'' of a static prompt, avoid unnecessary dynamic correction for easy samples, and reduce computational cost and noise.
\end{enumerate}

\subsection{Strategy for Selecting the Gating Threshold \texorpdfstring{$T$}{T}}
In the UCPOF framework, the gating threshold $T$ directly determines the system behavior. Rather than using a fixed hyperparameter, we set $T$ by choosing an error-coverage target based on a cost--benefit analysis on the calibration (sample) set.

\textbf{Generalization from training to test.} The core question is whether the threshold $T$ derived from the sample set can be applied to an unseen test set. We assume that the uncertainty distribution remains relatively stable across splits within the same domain. By targeting a high recall rate (90\%) on the sample set, we establish a safety margin for the system. Even under mild distribution shift on the test set, this conservative threshold prioritizes robust error detection over small fluctuations in retrieval efficiency.

\begin{figure}[t]
  \centering
  \includegraphics[width=0.92\linewidth]{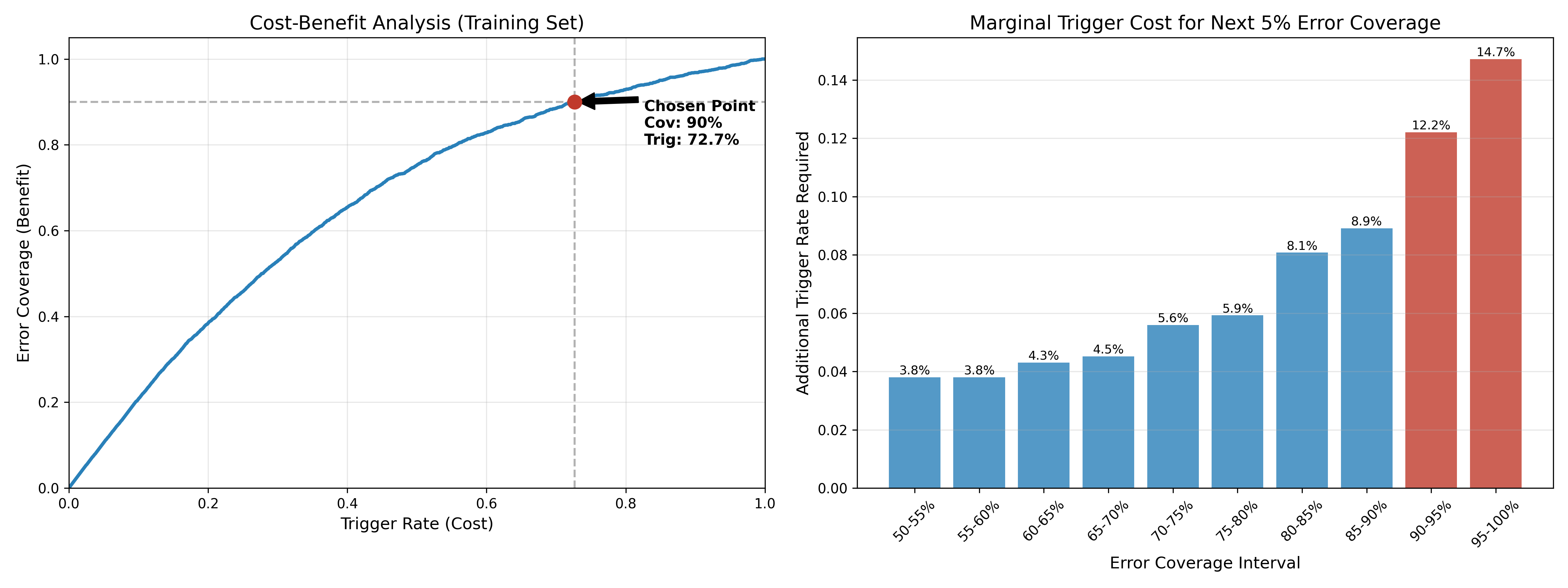}
  \caption{Relationship between error coverage and RAG trigger rate for threshold selection.}
  \label{fig:threshold_curve}
\end{figure}

As shown in Fig.~\ref{fig:threshold_curve}, we plot the relationship between error coverage (benefit) and RAG trigger rate (cost), which exhibits diminishing marginal returns. When the coverage rate is <85\%, the curve is steep, which means that a small amount of additional retrieval cost can be exchanged for a significant improvement in error coverage, and stopping at this time will not fully utilize the potential of RAG. When the coverage rate is >95\%, the curve tends to be flat, and the marginal cost surges. To capture the last 5\% of ``overconfidence errors'' (outliers, corresponding to the bottom edge of the ``peak'' in Fig.~\ref{fig:distribution_density}), an extremely aggressive threshold is required, resulting in a surge in the false trigger rate for correct samples. The 90\% coverage rate is exactly at the ``elbow'' of the curve, that is, the inflection point where the marginal utility begins to decline. Therefore, choosing a threshold that covers 90\% of the sample set errors represents the best statistical trade-off between efficiency and accuracy. Based on the assumption that the uncertainty distribution is relatively stable in the same domain (IID), the safety margin established on the sample set can ensure that the framework still has strong robustness when the test set distribution is slightly shifted.

\section{Experiments}
In this section, we verify the effectiveness of LSFU-driven prompt optimization through multi-dimensional experiments, focusing on the quality of static prompt generation, dynamic prompt correction, and the cost-effectiveness of the overall optimization framework. All experiments are designed around the core evaluation goal of ``prompt optimization effectiveness'' and cover task scenarios across different domains, difficulty levels, and distributional characteristics.

\subsection{Experimental Setup}
\textbf{Datasets}: We selected six datasets covering different domains, difficulty levels, and distributional characteristics: ACE and CASIE (event extraction with significant long-tail characteristics), AgNews (long-tail news classification), SST5 (fine-grained sentiment analysis), CB (natural language inference), and Reuters-21578 (R8) (long-tail text classification).

\textbf{Models}: We evaluate three models: Qwen2.5-7B-Instruct (main experimental model), Llama-3-8B-Instruct (cross-model generalization; Appendix A), and Qwen2.5-14B-Instruct (scalability with model size; Appendix A). All models use the default inference parameters (temperature = 0.1; maximum generated tokens = 10) to ensure that the experiments focus on prompt optimization rather than parameter tuning.

\textbf{Evaluation Metrics}: Accuracy, AUC for distinguishing correct vs. incorrect predictions, and average inference cost.

\textbf{Baselines}: Baseline (fixed prompt with randomly selected exemplars), Gold Shot (LSFU-based selection of low-uncertainty samples to construct a few-shot prompt), and Full RAG (always-on retrieval and correction for all test samples).

\subsection{Main Results}
Following Cobbina et al.~\citep{cobbina2025show}, we first examine how the ordering of prompt components affects guidance. Comparing the traditional structure of ``Task→Example→Input'' with the optimized structure of ``Example→Task→Input'', Table~\ref{tab:prompt_order} shows that the latter improves average accuracy by 2.79\%, which can reduce the model's cognitive load and strengthen example-based guidance. Therefore, all subsequent experiments (including Baseline, Gold Shot, and the static/reflective prompts in UCPOF) adopt the ``Example→Task→Input'' structure to ensure prompt-format consistency and avoid confounding factors. Experiments under the traditional ``Task→Example→Input'' structure are reported in Appendix B.

\begin{table}[t]
\centering
\caption{Impact of Prompt Component Ordering on Accuracy \%}
\label{tab:prompt_order}
\begin{tabular}{@{}lccccccc@{}}
\toprule
method & ACE & Casie & AgNews & SST5 & CB & R8 & AVG \\
\midrule
Task$\rightarrow$Example$\rightarrow$Input & 42.81 & 82.13 & 78.80 & 50.50 & 69.64 & 91.73 & 69.27 \\
Example$\rightarrow$Task$\rightarrow$Input & 51.07 & 76.40 & 82.42 & 51.45 & 78.57 & 92.46 & 72.06 \\
\bottomrule
\end{tabular}
\end{table}

\begin{table}[t]
\centering
\caption{Accuracy Comparison Experiment (Accuracy \%)}
\label{tab:main_results}
\begin{tabular}{@{}lccccccc@{}}
\toprule
method & ACE & Casie & AgNews & SST5 & CB & R8 & AVG \\
\midrule
baseline & 51.07 & 76.40 & 82.42 & 51.45 & 78.57 & 92.46 & 72.06 \\
gold shot (Ours) & 50.67 & 76.73 & 84.90 & 51.90 & 83.93 & 94.75 & 73.81 \\
Full RAG & 61.47 & 84.20 & 91.42 & 51.40 & 48.21 & 97.35 & 72.34 \\
UCPOF (Ours) & 62.08 & 85.40 & 89.99 & 51.54 & 82.14 & 97.40 & 78.09 \\
\bottomrule
\end{tabular}
\end{table}

\textbf{Analysis of Table~\ref{tab:main_results}.}
\begin{enumerate}
  \item \textbf{Overall gains.} UCPOF consistently improves over the static Baseline across all datasets, achieving a +6.03\% average accuracy increase. The largest gain appears on ACE (+11.01\%), indicating that uncertainty-calibrated, dynamic intervention is particularly beneficial on difficult, ambiguity-prone tasks.
  \item \textbf{UCPOF vs\. Full RAG: retrieval as a gated correction rather than a default.} On ACE and Casie, UCPOF clearly outperforms Full RAG, suggesting that always-on retrieval can inject irrelevant or misleading context for many samples. Notably, Full RAG fails on CB (48.21\% vs.\ 78.57\% for Baseline), whereas UCPOF avoids this collapse by retrieving only when the first-token uncertainty indicates risk. On AgNews, Full RAG is slightly higher than UCPOF (91.42\% vs.\ 89.99\%), which we attribute to the dataset being robust to retrieval noise; in such cases UCPOF can still be preferable due to its substantially lower retrieval rate and thus lower computation.
  \item \textbf{Why purely static exemplar selection is insufficient.} Gold Shot improves the average accuracy but can underperform Baseline on challenging datasets (e.g., ACE: 51.07\% $\rightarrow$ 50.67\%). A plausible explanation is that selecting only low-uncertainty exemplars over-emphasizes easy patterns and reduces coverage of hard cases. UCPOF addresses this limitation by using uncertainty to trigger retrieval and by dynamically correcting prompts at inference time.
\end{enumerate}

Overall, UCPOF achieves the best average performance, while the LSFU-based Gold Shot strategy provides a strong static baseline, jointly supporting the effectiveness of LSFU and uncertainty-gated dynamic retrieval.

\subsection{Ablation Studies}
\subsubsection{Gold Shot Selection Strategy}
To validate the rationale of using ``LSFU-screened low-uncertainty samples'' as static prompt exemplars, we compare three selection strategies (Table~\ref{tab:gold_shot_ablation}): random selection (Random; used in Baseline), selecting the hardest samples (Highest LSFU; simulating how borderline examples can mislead the model), and selecting the easiest samples (Lowest LSFU/Ours; our gold-shot screening strategy).

\begin{table}[t]
\centering
\caption{Gold Shot Selection Strategy Comparison (Stage 1 Accuracy \%)}
\label{tab:gold_shot_ablation}
\begin{tabular}{@{}lccccccc@{}}
\toprule
Strategy & ACE & Casie & AgNews & SST5 & CB & R8 & AVG \\
\midrule
Random & 51.07 & 76.40 & 82.42 & 51.45 & 78.57 & 92.46 & 72.06 \\
Highest LSFU & 45.87 & 74.87 & 87.62 & 53.98 & 69.64 & 90.95 & 70.49 \\
Lowest LSFU (ours) & 50.67 & 76.73 & 84.90 & 51.90 & 83.93 & 94.75 & 73.81 \\
\bottomrule
\end{tabular}
\end{table}

Using the ``hardest'' samples (Highest LSFU) as exemplars substantially reduces accuracy on the most challenging datasets (ACE, CB, R8). This supports the intuition that boundary cases are unstable anchors: they are close to the decision boundary and may increase the model's reasoning burden in few-shot prompting. In contrast, the ``easiest'' samples (Lowest LSFU) provide stable, prototypical patterns and yield the best average accuracy (73.81\%).

We further evaluate \emph{UCPOF w/o Gold Shot}, where the static prompt uses random exemplars and only the dynamic correction stage is retained (Table~\ref{tab:gold_shot_ablation2}), to quantify the contribution of Gold Shot to the overall framework.

\begin{table}[t]
\centering
\caption{Ablation of Gold Shot Selection in UCPOF}
\label{tab:gold_shot_ablation2}
\begin{tabular}{@{}lccccccc@{}}
\toprule
Strategy & ACE & Casie & AgNews & SST5 & CB & R8 & AVG \\
\midrule
UCPOF & 62.08 & 85.40 & 89.99 & 51.54 & 82.14 & 97.40 & 78.09 \\
w/o Gold Shot & 61.47 & 84.93 & 89.08 & 51.53 & 71.43 & 96.90 & 75.89 \\
\bottomrule
\end{tabular}
\end{table}

Removing Gold Shot decreases the average accuracy by 2.20\% (78.09\% $\rightarrow$ 75.89\%), confirming that high-quality static exemplars are an important foundation for effective dynamic correction.

\subsubsection{Necessity of the Prior Probability}
To verify the core role of $P_{\mathrm{prior}}$ in the LSFU formula, we compare the AUC index of LSFU and ordinary First Token Entropy ($\log_{10}(H(P_{\mathrm{top}\text{-}k}(x))+\varepsilon)$) in distinguishing correct and incorrect predictions (LSFU(with prior) vs Entropy (without prior), Table~\ref{tab:prior_ablation_auc}) and the accuracy of the corresponding prompt strategy (Table~\ref{tab:prior_ablation_acc}).

\begin{table}[t]
\centering
\caption{Ablation of $P_{\mathrm{prior}}$ (AUC Metric)}
\label{tab:prior_ablation_auc}
\begin{tabular}{@{}llccccccc@{}}
\toprule
model & & ACE & Casie & AgNews & SST5 & CB & R8 & AVG \\
\midrule
Main model & & 0.8104 & 0.7492 & 0.7761 & 0.5242 & 0.7568 & 0.9161 & 0.7555 \\
& w/o $P_{\mathrm{prior}}$ & 0.8004 & 0.7338 & 0.7762 & 0.5235 & 0.7154 & 0.9083 & 0.7429 \\
\bottomrule
\end{tabular}
\end{table}

On Qwen2.5-7B-Instruct, introducing $P_{\mathrm{prior}}$ consistently improves AUC across datasets, indicating that accounting for label priors helps separate prior-dominated confidence from context-driven certainty. Additional cross-model ablations are provided in Appendix~\ref{app:prior_ablation_other_models}.

\begin{table}[t]
\centering
\caption{Ablation of $P_{\mathrm{prior}}$ in UCPOF (ACC Metric)}
\label{tab:prior_ablation_acc}
\begin{tabular}{@{}llccccccc@{}}
\toprule
Method & & ACE & Casie & AgNews & SST5 & CB & R8 & AVG \\
\midrule
UCPOF & & 62.08 & 85.40 & 89.99 & 51.54 & 82.14 & 97.40 & 78.09 \\
& w/o $P_{\mathrm{prior}}$ & 61.39 & 84.80 & 89.08 & 52.44 & 60.71 & 97.85 & 74.37 \\
\bottomrule
\end{tabular}
\end{table}

\begin{table}[t]
\centering
\caption{Ablation of $P_{\mathrm{prior}}$ (Calibration Metrics; lower is better)}
\label{tab:prior_ablation_calibration}
\begin{tabular}{@{}llccccccc@{}}
\toprule
Metric & & ACE & Casie & AgNews & SST5 & CB & R8 & AVG \\
\midrule
ECE & & 0.2240 & 0.1354 & 0.0978 & 0.3545 & 0.1245 & 0.1422 & 0.1797 \\
& w/o $P_{\mathrm{prior}}$ & 0.2271 & 0.1365 & 0.1015 & 0.3547 & 0.2046 & 0.1421 & 0.1944 \\
Brier score & & 0.2604 & 0.1368 & 0.0978 & 0.3896 & 0.1246 & 0.1431 & 0.1920 \\
& w/o $P_{\mathrm{prior}}$ & 0.2634 & 0.1374 & 0.1023 & 0.3896 & 0.2031 & 0.1426 & 0.2064 \\
\bottomrule
\end{tabular}
\end{table}

In addition to AUC and accuracy, we report calibration metrics (Table~\ref{tab:prior_ablation_calibration}): \emph{Expected Calibration Error} (ECE), which measures the gap between predicted confidence and empirical accuracy via binning, and the \emph{Brier score}, which measures the mean squared error between predicted probabilities and the one-hot labels. Lower values indicate better calibration. Both ECE and Brier score decrease after incorporating $P_{\mathrm{prior}}$, indicating improved confidence calibration. The largest improvement appears on CB, suggesting that prior-aware uncertainty is particularly effective at mitigating miscalibration on challenging, long-tail-like settings.

\subsection{Efficiency Analysis}
To evaluate the cost-effectiveness of prompt optimization---i.e., whether UCPOF improves accuracy while reducing unnecessary computation---we conduct two analyses.

First, we plot an \emph{efficiency--accuracy} Pareto curve, where the x-axis is average token consumption (static prompt tokens plus dynamically retrieved tokens) and the y-axis is accuracy. We compare UCPOF with Baseline and Full RAG to test whether UCPOF achieves a better accuracy--efficiency trade-off.

\begin{figure}[t]
  \centering
  \includegraphics[width=0.92\linewidth]{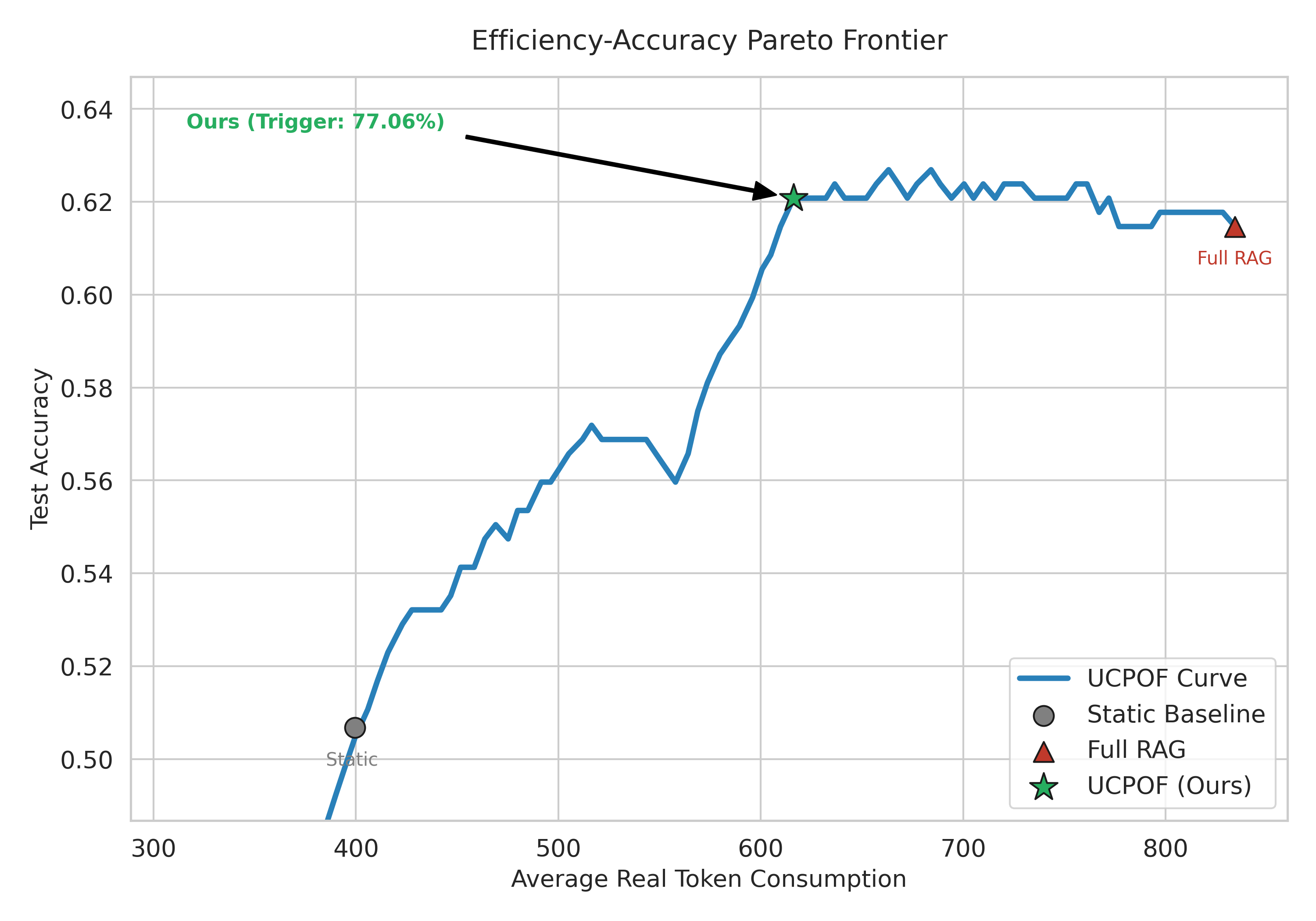}
  \caption{Efficiency--Accuracy Pareto Frontier (ACE; Qwen2.5-7B-Instruct).}
  \label{fig:pareto_curve}
\end{figure}

\textbf{Pareto optimality (Fig.~\ref{fig:pareto_curve}).} The green star denotes UCPOF's operating point on the test set (using the 90\% error-coverage threshold selected on the sample set). The gray dot labeled ``static'' denotes UCPOF's initial \emph{static-prompt} state before any retrieval is triggered; it corresponds to the \emph{Gold Shot} static prompt (Table~2), where inference is performed using only the offline-selected gold-shot exemplars without RAG. UCPOF lies near the upper-left region of the Pareto frontier, outperforming Baseline (lower-left) and Full RAG (upper-right). We also observe a drop near the high-cost end of the frontier (close to Full RAG), suggesting that forcing retrieval on the remaining hard tail can introduce harmful noise and reduce accuracy. UCPOF stops before this regime, achieving a favorable cost--performance balance.

Second, we plot a \emph{retrieval-trigger-rate vs.\ accuracy} curve, where the trigger rate is the fraction of samples for which LSFU activates dynamic correction. This analysis checks whether the offline-selected 90\% error-coverage threshold lies in an efficient regime that achieves high accuracy without excessive retrieval.

\begin{figure}[t]
  \centering
  \includegraphics[width=0.92\linewidth]{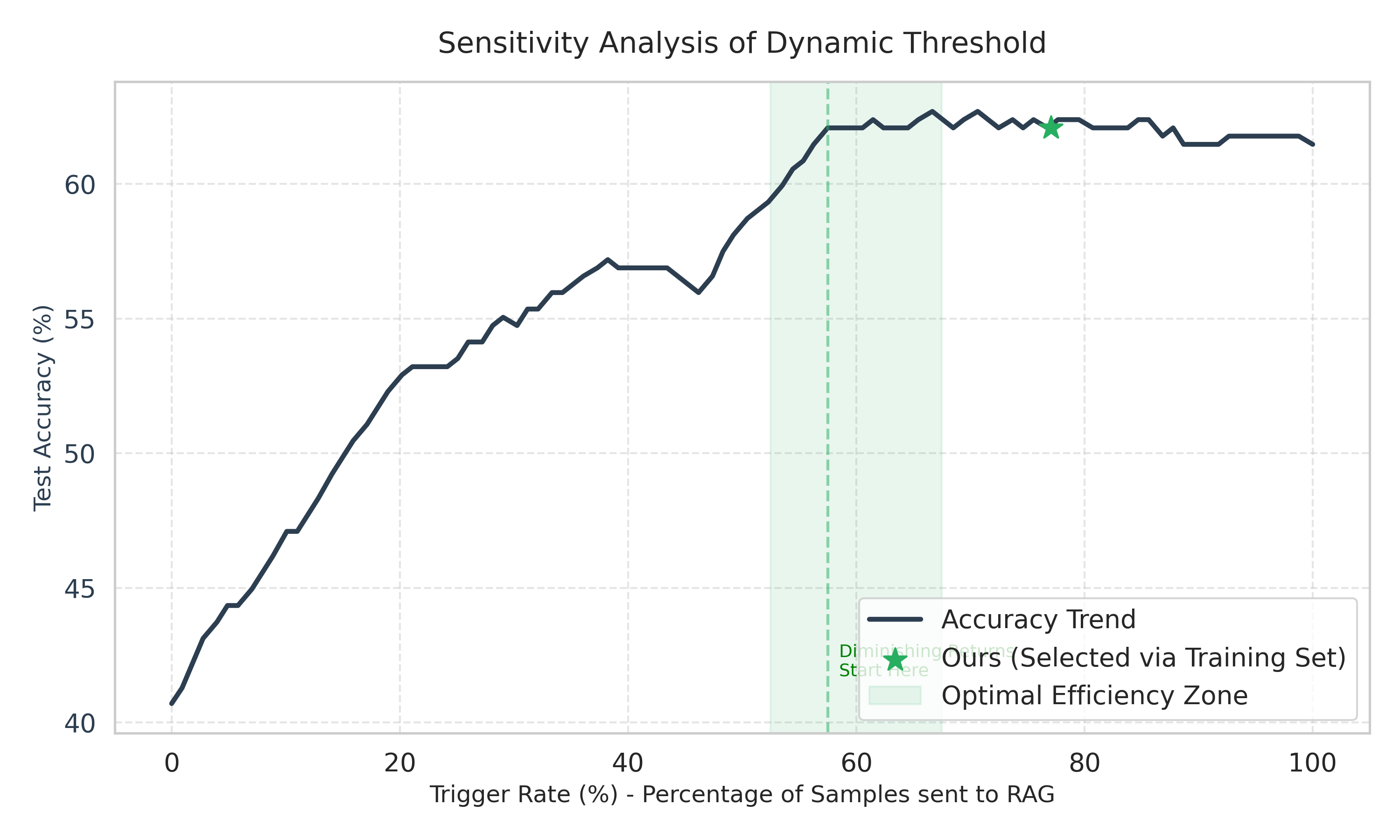}
  \caption{RAG Trigger Rate vs.\ Test Accuracy Trend (ACE; Qwen2.5-7B-Instruct).}
  \label{fig:trigger_rate_curve}
\end{figure}

\textbf{Threshold robustness (Fig.~\ref{fig:trigger_rate_curve}).} Accuracy saturates once the trigger rate reaches roughly 60\%. The threshold tuned on the sample set yields a trigger rate of $\sim$77\% on the test set (green star). Although this is beyond the inflection point, it reflects a deliberate safety margin: we spend modest additional compute in the saturation region to better capture long-tail errors and reduce the risk of missing difficult samples due to an overly conservative threshold.

\paragraph{Efficiency comparison with Full RAG.}
To highlight UCPOF's practical efficiency advantage over always-on retrieval, we further compare UCPOF with Full RAG in terms of (i) \emph{retrieval trigger rate} (the fraction of test samples that invoke retrieval) and (ii) \emph{test accuracy}. As shown in Table~\ref{tab:ucpof_vs_fullrag_efficiency}, Full RAG always retrieves (trigger rate = 100\%), whereas UCPOF substantially reduces the trigger rate across most datasets while maintaining or improving accuracy. In particular, UCPOF improves the overall average accuracy by +5.75 points (78.09 vs.\ 72.34) while cutting the average retrieval trigger rate by 50.66\% (49.34\% vs.\ 100\%), demonstrating that selective retrieval can both reduce retrieval-induced noise and lower computational cost.

\begin{table}[t]
\centering
\caption{Efficiency comparison between UCPOF and Full RAG: retrieval trigger rate (lower is better) and test accuracy (higher is better).}
\label{tab:ucpof_vs_fullrag_efficiency}
\begin{tabular}{@{}llccccccc@{}}
\toprule
Method & Metric & ACE & Casie & AgNews & SST5 & CB & R8 & AVG \\
\midrule
\multirow{2}{*}{UCPOF} & Trigger rate (\%) & 77.06 & 50.60 & 20.87 & 95.48 & 16.07 & 35.95 & 49.34 \\
& Acc & 62.08 & 85.40 & 89.99 & 51.54 & 82.14 & 97.40 & 78.09 \\
\midrule
\multirow{2}{*}{Full RAG} & Trigger rate (\%) & 100.00 & 100.00 & 100.00 & 100.00 & 100.00 & 100.00 & 100.00 \\
& Acc & 61.47 & 84.20 & 91.42 & 51.40 & 48.21 & 97.35 & 72.34 \\
\bottomrule
\end{tabular}
\end{table}

\paragraph{Experimental setting for Fig.~\ref{fig:pareto_curve}--\ref{fig:trigger_rate_curve}.}
Unless otherwise specified, the efficiency analyses in this subsection are conducted on the \textbf{ACE} dataset using \textbf{Qwen2.5-7B-Instruct}. For completeness, we provide the corresponding Pareto and trigger-rate curves for the other datasets in Appendix~\ref{app:efficiency_other_datasets}.

\section{Discussion and Conclusion}
Aiming at the core problems of prompt design sensitivity, uncertainty measurement distortion and full RAG noise interference in in-context learning (ICL), this paper proposes a prior-aware log-scale focal uncertainty (LSFU) metric and a prompt optimization framework (UCPOF), which provides an efficient and reliable inference paradigm for classification tasks. The main conclusions are as follows:

1. The LSFU metric achieves more accurate confidence calibration and provides a quantitative basis for prompt optimization. This metric combines the first-token semantic features and category prior distribution, effectively distinguishes ``spurious confidence'' from ``true certainty'' through nonlinear risk weighting, and improves the average AUC index across models and datasets by 1.26\%, solving the prior bias defect of traditional entropy.

2. The LSFU-driven gold example screening strategy optimizes static prompt generation. Selecting high-confidence samples with low LSFU values as Few-shot examples improves the average accuracy by 1.75\% compared with random selection, providing clear decision anchors for the model, and reducing cognitive noise and hallucination propagation risks.

3. The UCPOF framework achieves a balance between the effect and efficiency of prompt optimization. Through the two-stage mechanism of ``static prompt optimization + dynamic prompt correction'', with LSFU as the gate to accurately trigger retrieval, the average accuracy reaches 78.09\% (6.03\% higher than Baseline), and the inference cost is lower than that of full RAG, effectively filtering redundant noise.

The main limitations of the research are reflected in three aspects: first, the task adaptation scope is limited, only designed for classification tasks, and there have been special studies on large language models in fine-grained classification tasks such as event extraction \citep{li2025event}, and the adaptability of this method can be expanded based on its research results in the future; second, the construction of the prompt does not fully optimize the example arrangement and instruction details; third, the generalization of the dynamic threshold depends on the stability of the domain data distribution.

Future work will be carried out around three directions: first, expand LSFU to unsupervised prompt generation to reduce the dependence on labeled data; second, optimize the prompt structure of complex tasks to improve the guidance accuracy; third, explore the application of LSFU in retriever re-ranking and multi-round prompt optimization, and adapt to more task types.

\bibliographystyle{plainnat} 
\bibliography{main} 

\appendix
\section{Efficiency Analysis on Other Datasets}
\label{app:efficiency_other_datasets}
This appendix reports the efficiency--accuracy Pareto curves and the retrieval-trigger-rate vs.\ accuracy curves on datasets other than ACE, using the same experimental protocol as in Sec.~6.4.

\begin{figure}[t]
  \centering
  \begin{subfigure}{0.49\textwidth}
    \centering
    \includegraphics[width=\linewidth]{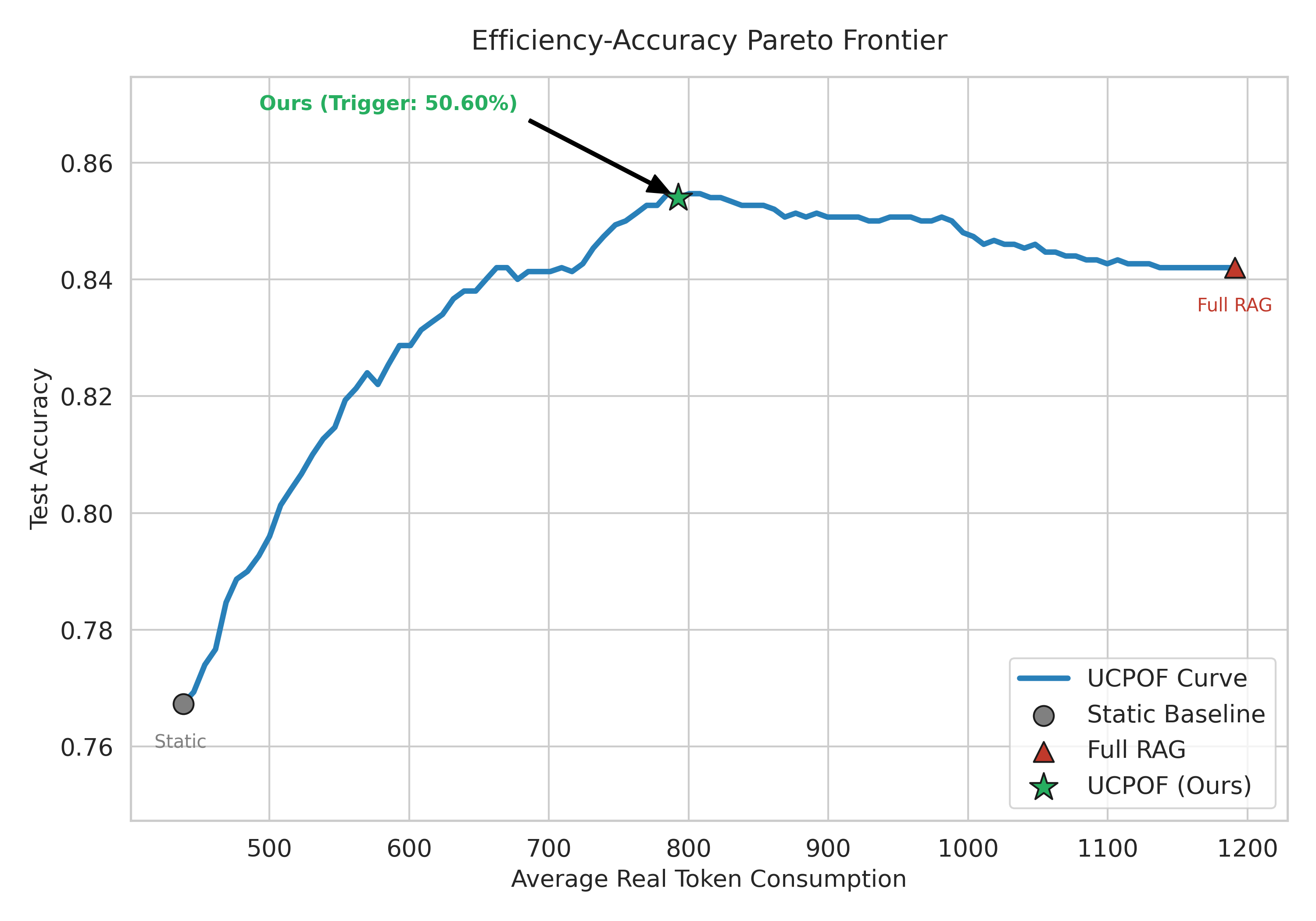}
    \caption{}
    \label{fig:pareto_curve_casie}
  \end{subfigure}
  \hfill
  \begin{subfigure}{0.49\textwidth}
    \centering
    \includegraphics[width=\linewidth]{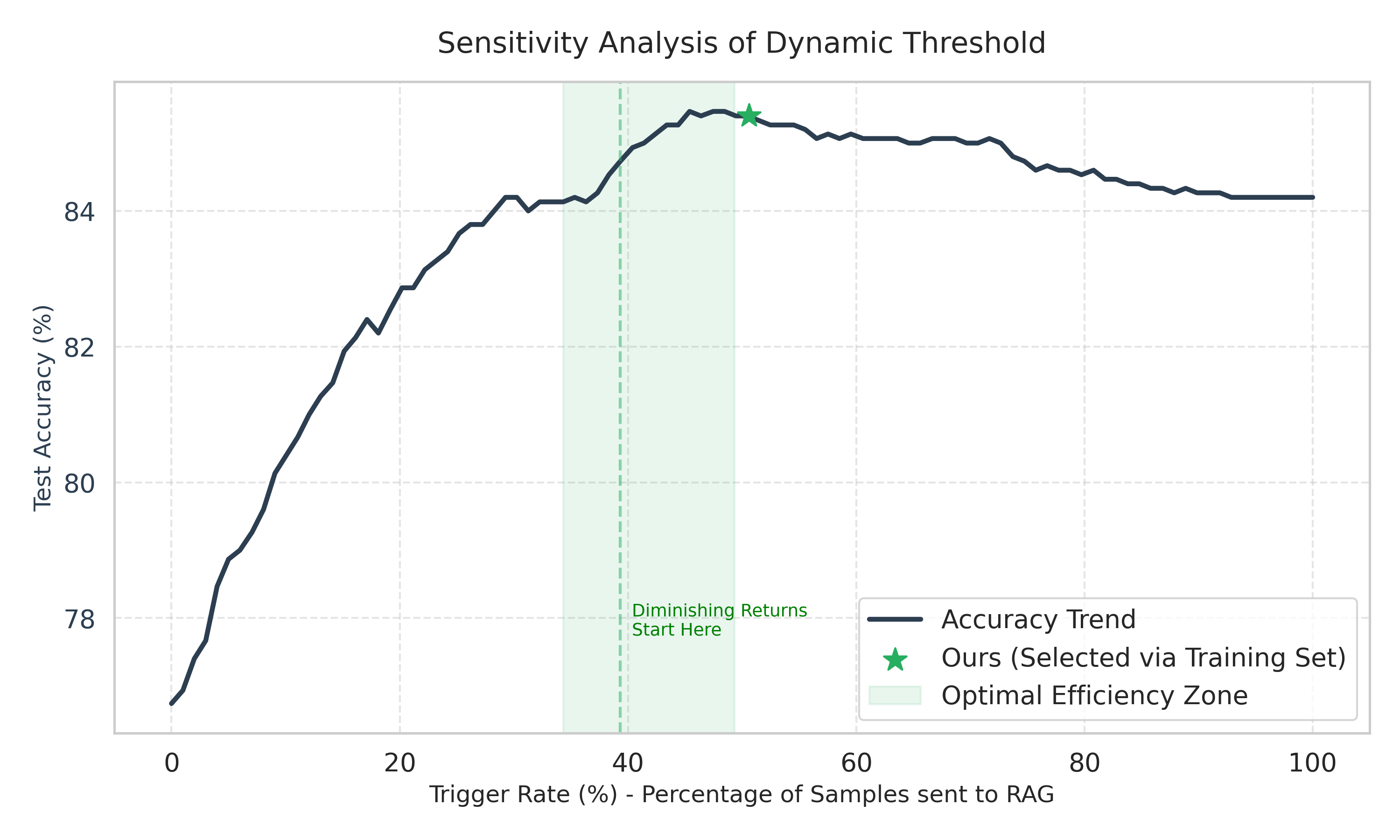}
    \caption{}
    \label{fig:trigger_rate_curve_casie}
  \end{subfigure}
  \caption{Efficiency analysis on CASIE.}
  \label{fig:efficiency_curves_casie}
\end{figure}

\begin{figure}[t]
  \centering
  \begin{subfigure}{0.49\textwidth}
    \centering
    \includegraphics[width=\linewidth]{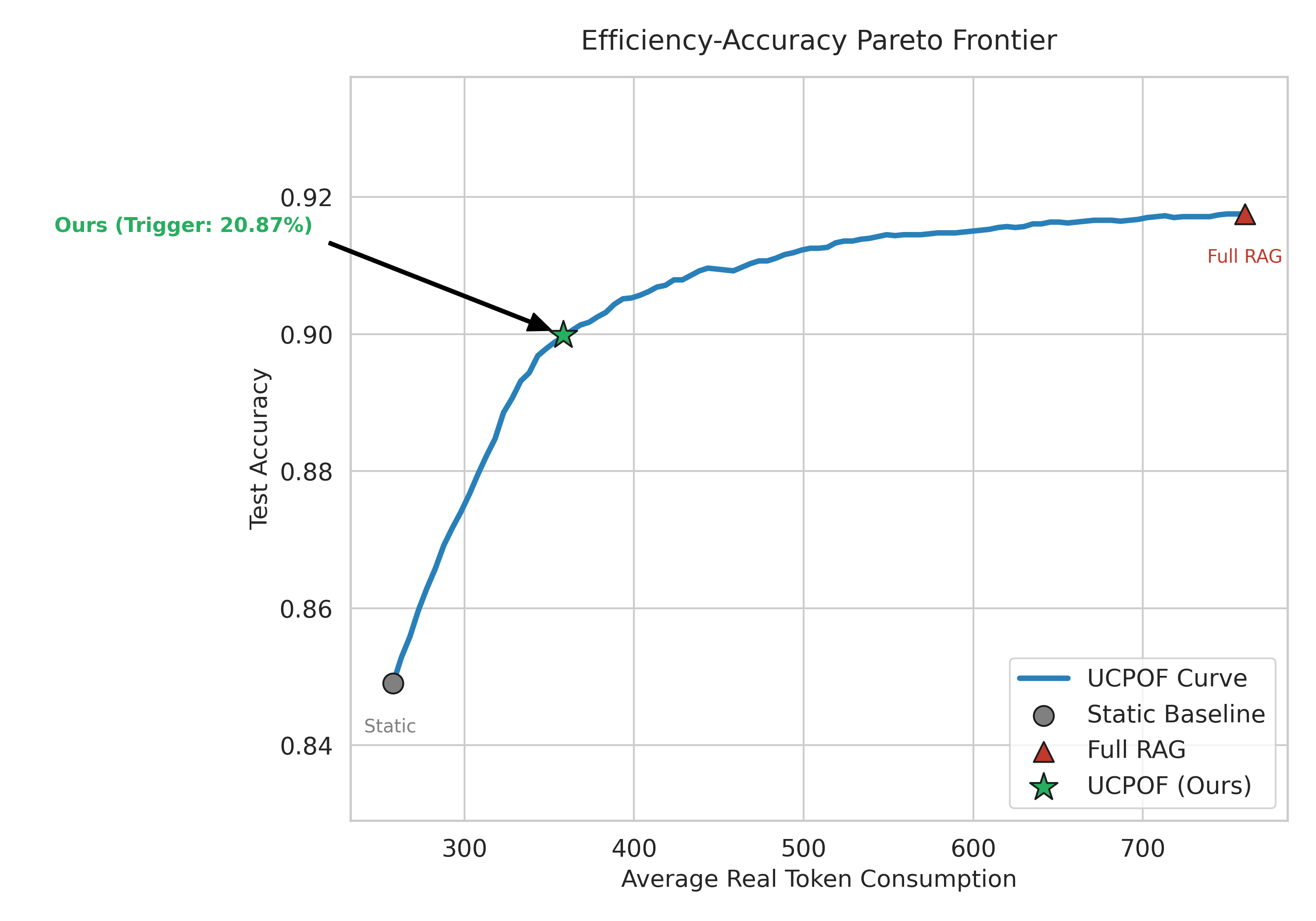}
    \caption{}
    \label{fig:pareto_curve_agnews}
  \end{subfigure}
  \hfill
  \begin{subfigure}{0.49\textwidth}
    \centering
    \includegraphics[width=\linewidth]{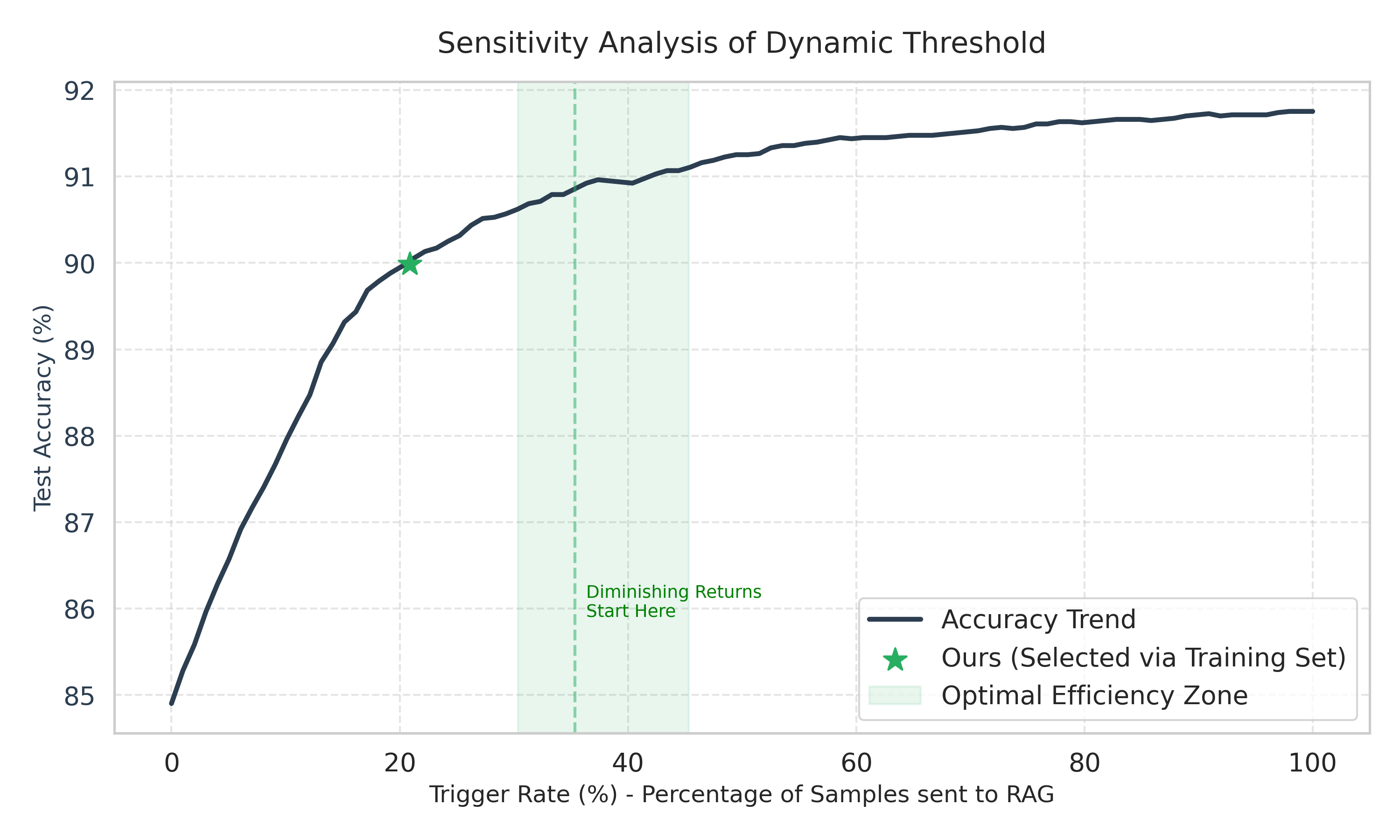}
    \caption{}
    \label{fig:trigger_rate_curve_agnews}
  \end{subfigure}
  \caption{Efficiency analysis on AgNews.}
  \label{fig:efficiency_curves_agnews}
\end{figure}

\begin{figure}[t]
  \centering
  \begin{subfigure}{0.49\textwidth}
    \centering
    \includegraphics[width=\linewidth]{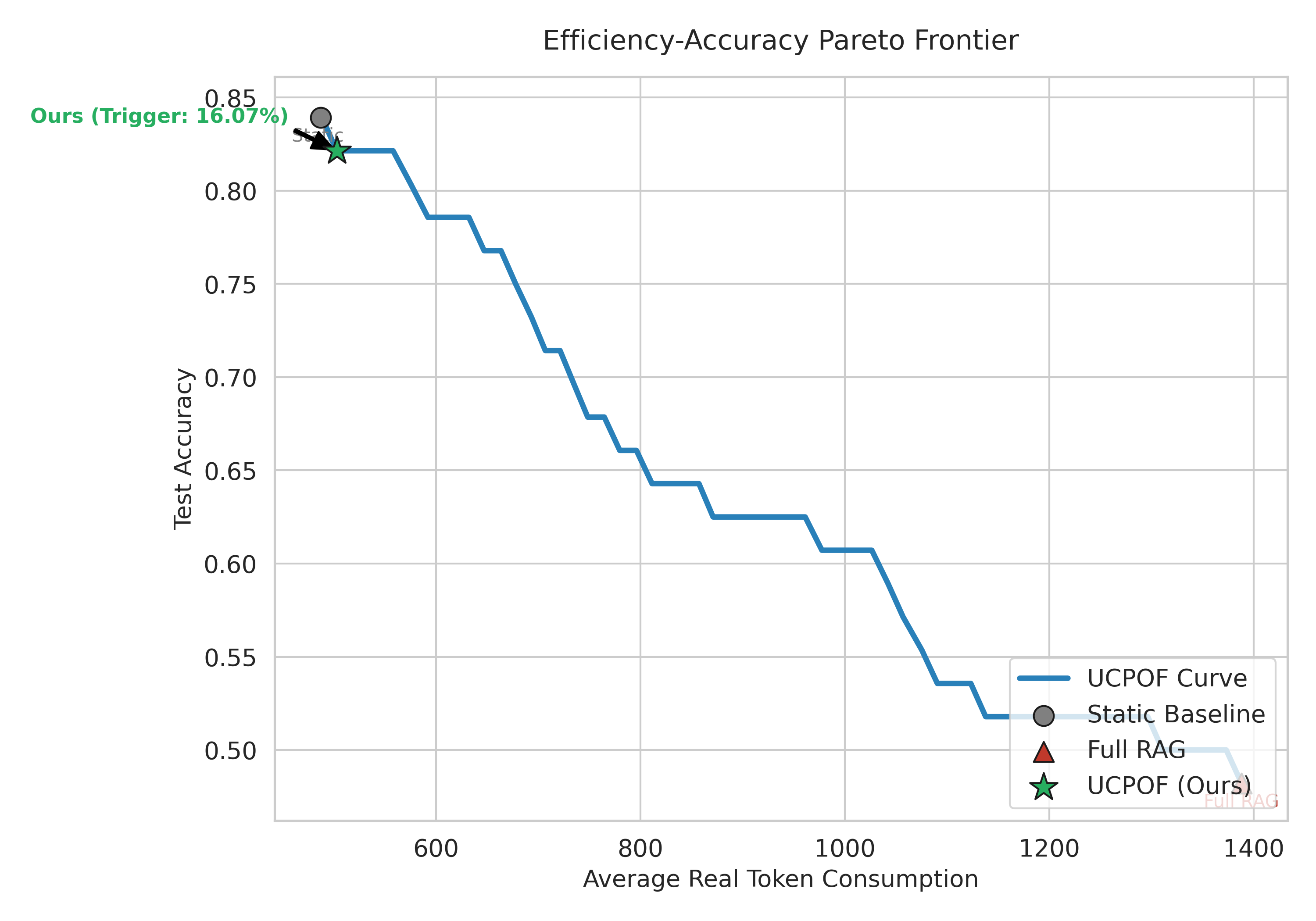}
    \caption{}
    \label{fig:pareto_curve_cb}
  \end{subfigure}
  \hfill
  \begin{subfigure}{0.49\textwidth}
    \centering
    \includegraphics[width=\linewidth]{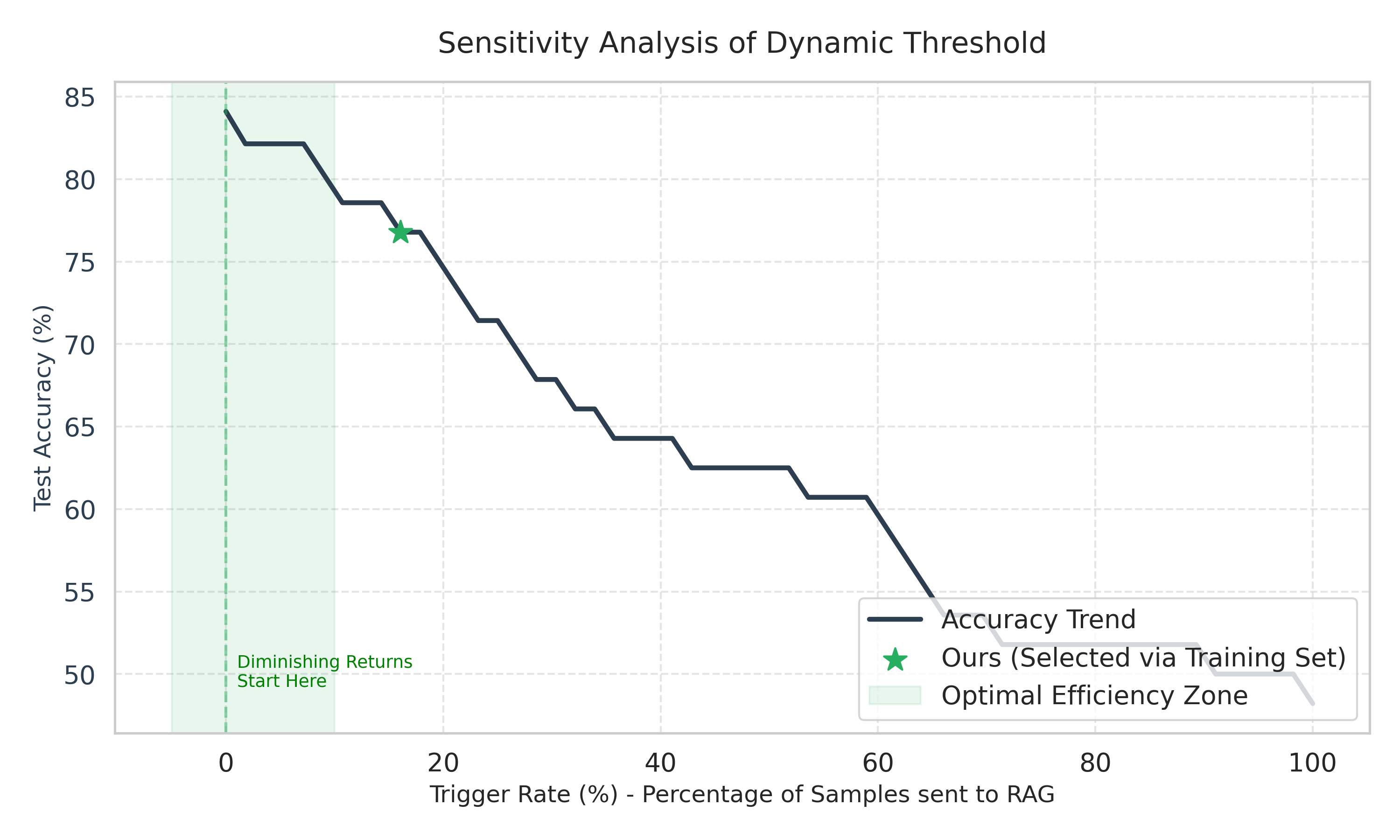}
    \caption{}
    \label{fig:trigger_rate_curve_cb}
  \end{subfigure}
  \caption{Efficiency analysis on CB.}
  \label{fig:efficiency_curves_cb}
\end{figure}

\begin{figure}[t]
  \centering
  \begin{subfigure}{0.49\textwidth}
    \centering
    \includegraphics[width=\linewidth]{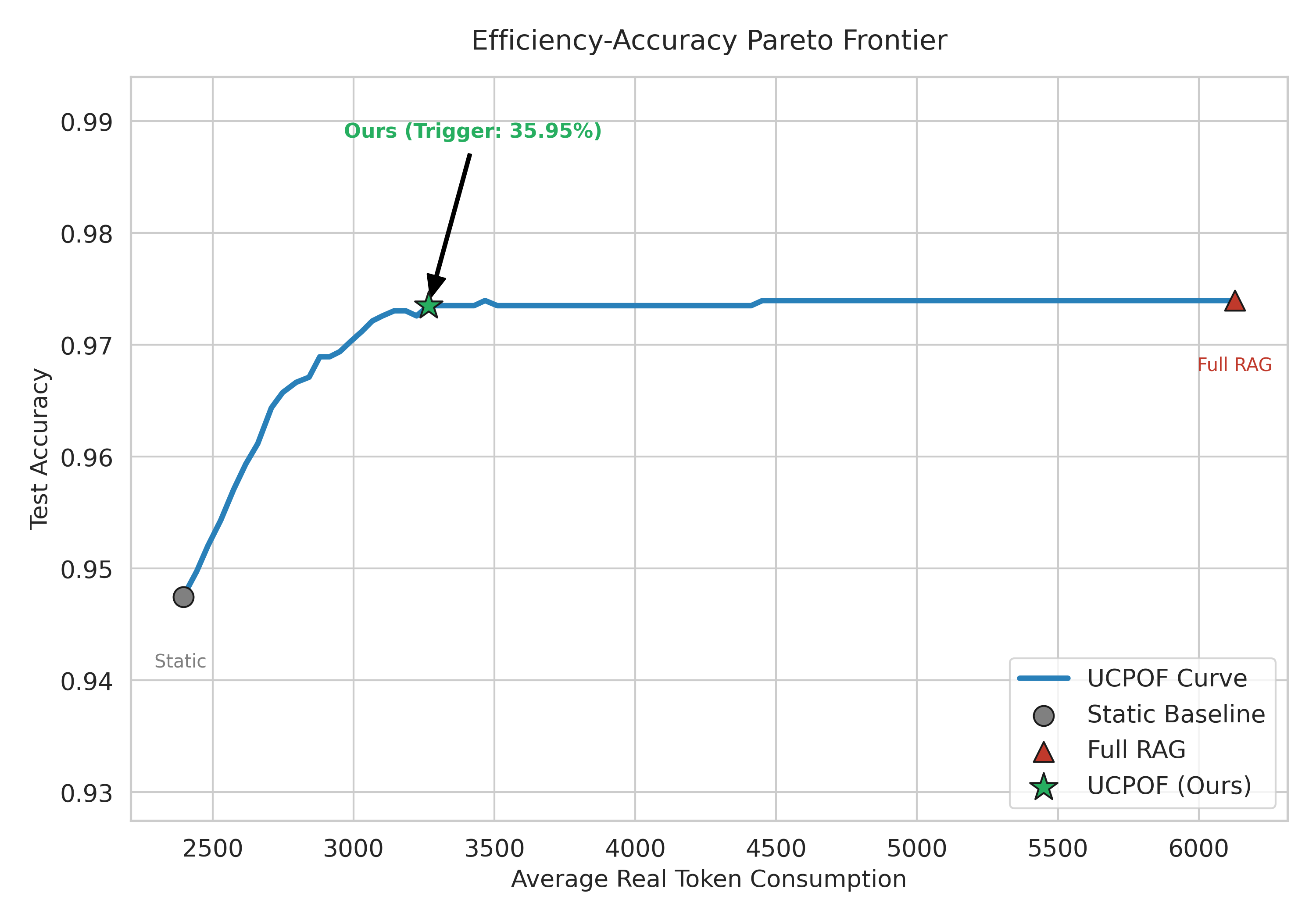}
    \caption{}
    \label{fig:pareto_curve_r8}
  \end{subfigure}
  \hfill
  \begin{subfigure}{0.49\textwidth}
    \centering
    \includegraphics[width=\linewidth]{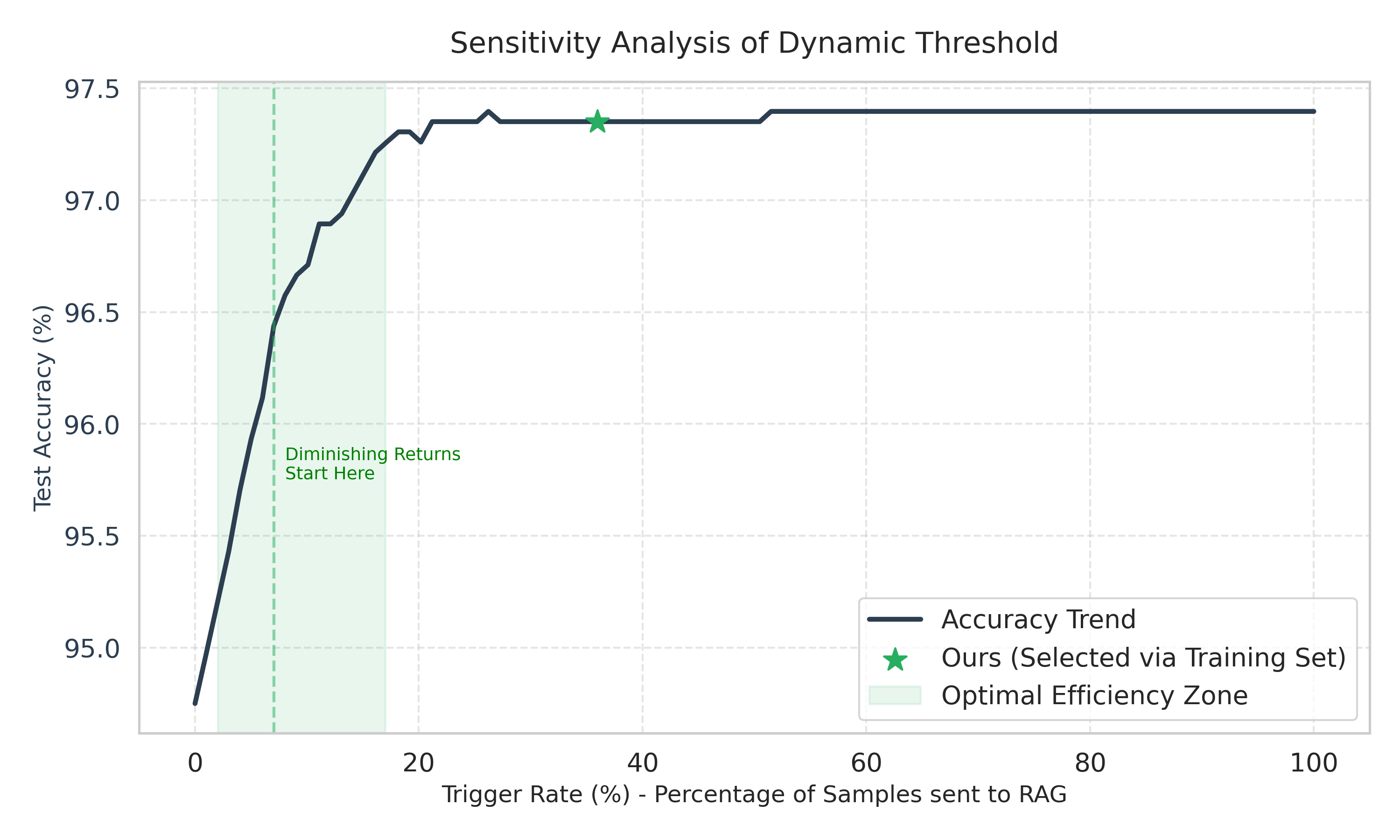}
    \caption{}
    \label{fig:trigger_rate_curve_r8}
  \end{subfigure}
  \caption{Efficiency analysis on R8.}
  \label{fig:efficiency_curves_r8}
\end{figure}

\section{Cross-Model Scalability}
To verify whether LSFU depends on a specific model architecture, we repeated the experiments on Llama-3 and Qwen models of different sizes.

\begin{table}[h]
\centering
\caption{Cross-Model Experimental Results - Llama-3-8B-Instruct}
\begin{tabular}{@{}lccccccc@{}}
\toprule
method & ACE & Casie & AgNews & SST5 & CB & R8 & AVG \\
\midrule
baseline & 33.94 & 62.73 & 80.47 & 48.10 & 64.29 & 93.19 & 63.79 \\
gold shot (Ours) & 37.61 & 65.53 & 82.24 & 46.52 & 62.50 & 93.51 & 64.65 \\
Full RAG & 60.55 & 82.73 & 90.92 & 50.63 & 82.14 & 96.03 & 73.83 \\
UCPOF (Ours) & 55.96 & 81.73 & 90.80 & 47.19 & 83.93 & 96.39 & 76.00 \\
\bottomrule
\end{tabular}
\end{table}

\begin{table}[h]
\centering
\caption{Cross-Model Experimental Results - Qwen2.5-14B-Instruct}
\begin{tabular}{@{}lccccccc@{}}
\toprule
method & ACE & Casie & AgNews & SST5 & CB & R8 & AVG \\
\midrule
baseline & 52.60 & 83.07 & 80.25 & 53.08 & 87.50 & 92.51 & 74.82 \\
gold shot (Ours) & 47.09 & 86.53 & 86.40 & 53.26 & 81.36 & 93.42 & 74.68 \\
Full RAG & 65.44 & 89.47 & 91.50 & 54.12 & 82.14 & 97.94 & 80.10 \\
UCPOF (Ours) & 66.97 & 89.33 & 89.87 & 54.21 & 83.93 & 97.99 & 80.38 \\
\bottomrule
\end{tabular}
\end{table}

The results show that: in cross-model scenarios, UCPOF (LSFU-driven prompt optimization) is better than static baseline, static optimization (Gold Shot) and full RAG, with an average accuracy improvement of 2.17\%-3.56\%, proving that the prior calibration logic of LSFU does not depend on a specific model pretraining paradigm; even when the model parameters increase (Qwen2.5-14B), UCPOF with LSFU can still maintain performance advantages (average accuracy 80.38\% vs full RAG 80.10\%). On low-parameter models (Llama-3-8B), UCPOF has the most significant improvement in CB task (semantic sensitive) (83.93\% vs full RAG 82.14\%), proving that the noise filtering mechanism of LSFU is more valuable for scenarios with weak model capabilities, further supporting the necessity of dynamic prompt correction.

\section{Prior-Ablation Results on Other Models}
\label{app:prior_ablation_other_models}
Table~\ref{tab:prior_ablation_auc_other_models} reports the ablation results of $P_{\mathrm{prior}}$ on additional model backbones.

\begin{table}[t]
\centering
\caption{Ablation of $P_{\mathrm{prior}}$ (AUC Metric) on other models.}
\label{tab:prior_ablation_auc_other_models}
\begin{tabular}{@{}llccccccc@{}}
\toprule
model & & ACE & Casie & AgNews & SST5 & CB & R8 & AVG \\
\midrule
Qwen2.5-14B-instruct & & 0.6742 & 0.7295 & 0.7763 & 0.5409 & 0.8528 & 0.9058 & 0.7466 \\
& w/o $P_{\mathrm{prior}}$ & 0.6666 & 0.7240 & 0.7229 & 0.5427 & 0.8364 & 0.8941 & 0.7311 \\
Llama-3-8B-Instruct & & 0.7138 & 0.7603 & 0.8083 & 0.4869 & 0.9271 & 0.7976 & 0.7490 \\
& w/o $P_{\mathrm{prior}}$ & 0.6183 & 0.7136 & 0.7876 & 0.4852 & 0.8954 & 0.7579 & 0.7097 \\
\bottomrule
\end{tabular}
\end{table}

\begin{table}[t]
\centering
\caption{Ablation of $P_{\mathrm{prior}}$ (AUC Metric) on other models.}
\label{tab:prior_ablation_auc_other_models}
\begin{tabular}{@{}llccccccc@{}}
\toprule
model & & ACE & Casie & AgNews & SST5 & CB & R8 & AVG \\
\midrule
Qwen2.5-14B-instruct & & 0.6742 & 0.7295 & 0.7763 & 0.5409 & 0.8528 & 0.9058 & 0.7466 \\
& w/o $P_{\mathrm{prior}}$ & 0.6666 & 0.7240 & 0.7229 & 0.5427 & 0.8364 & 0.8941 & 0.7311 \\
Llama-3-8B-Instruct & & 0.7138 & 0.7603 & 0.8083 & 0.4869 & 0.9271 & 0.7976 & 0.7490 \\
& w/o $P_{\mathrm{prior}}$ & 0.6183 & 0.7136 & 0.7876 & 0.4852 & 0.8954 & 0.7579 & 0.7097 \\
\bottomrule
\end{tabular}
\end{table}

\section{Additional Experiments on Prompt Structure Validation}
To support Section 5.2 of the main text, we provide supplementary data across models and parameters under the traditional structure of ``Task→Example→Input'' to strengthen the reliability of prompt optimization.

\begin{table}[h]
\centering
\caption{Cross-Model Experimental Results-Qwen2.5-7B-Instruct (Task→Example→Input)}
\begin{tabular}{@{}lccccccc@{}}
\toprule
method & ACE & Casie & AgNews & SST5 & CB & R8 & AVG \\
\midrule
baseline & 42.81 & 82.13 & 78.80 & 50.50 & 69.64 & 91.73 & 69.27 \\
gold shot (Ours) & 42.81 & 76.20 & 85.27 & 53.26 & 73.21 & 92.69 & 70.57 \\
Full RAG & 61.77 & 83.47 & 91.71 & 52.26 & 50.00 & 96.67 & 72.65 \\
UCPOF (Ours) & 62.69 & 84.33 & 89.72 & 52.35 & 70.86 & 95.84 & 75.97 \\
\bottomrule
\end{tabular}
\end{table}

\begin{table}[h]
\centering
\caption{Cross-Model Experimental Results-Qwen2.5-14B-Instruct (Task→Example→Input)}
\begin{tabular}{@{}lccccccc@{}}
\toprule
method & ACE & Casie & AgNews & SST5 & CB & R8 & AVG \\
\midrule
baseline & 49.54 & 83.53 & 83.72 & 55.61 & 87.50 & 95.29 & 75.87 \\
gold shot (Ours) & 42.53 & 87.40 & 84.98 & 51.90 & 80.36 & 95.80 & 73.82 \\
Full RAG & 67.58 & 90.00 & 91.75 & 53.67 & 85.71 & 97.99 & 81.12 \\
UCPOF (Ours) & 67.58 & 89.93 & 88.27 & 53.98 & 89.29 & 97.99 & 81.17 \\
\bottomrule
\end{tabular}
\end{table}

\begin{table}[h]
\centering
\caption{Cross-Model Experimental Results-LLama-3-8B-Instruct (Task→Example→Input)}
\begin{tabular}{@{}lccccccc@{}}
\toprule
method & ACE & Casie & AgNews & SST5 & CB & R8 & AVG \\
\midrule
baseline & 41.28 & 80.33 & 77.15 & 44.89 & 53.21 & 95.34 & 65.37 \\
gold shot (Ours) & 39.49 & 73.27 & 80.16 & 48.51 & 67.86 & 89.17 & 66.41 \\
Full RAG & 57.19 & 82.53 & 90.71 & 51.72 & 82.14 & 91.41 & 75.95 \\
UCPOF (Ours) & 57.79 & 81.16 & 88.68 & 51.72 & 82.14 & 96.16 & 76.27 \\
\bottomrule
\end{tabular}
\end{table}

\section{Validating the Top-{K} Parameter in LSFU}
To support the claim in Section~3.1 that $K=50$ is an appropriate choice for computing LSFU, we evaluate the stability of the top-$K$ entropy term under different $K$ values. Specifically, we compute the average entropy for $K \in\{50,100,500,\text{ALL}\ \}$ and report the results in Table~\ref{tab:k_validation}. The differences across $K$ are below 0.01, indicating that $K=50$ already captures the relevant high-probability mass (including frequent distractors) and that increasing $K$ yields negligible benefit while incurring additional cost.

\begin{table}[h] 
\centering
\caption{Top-K Entropy Values with Different K (LSFU Calculation Stability Validation)}
\label{tab:k_validation}
\begin{tabular}{@{}lcccccc@{}}
\toprule
K & ACE & Casie & AgNews & SST5 & CB & R8 \\
\midrule
50 & 0.443084 & 0.187070 & 0.026404 & 0.188524 & 0.104495 & 0.042302 \\
100 & 0.451381 & 0.187354 & 0.026403 & 0.188524 & 0.104495 & 0.042302 \\
500 & 0.451371 & 0.187555 & 0.026412 & 0.188525 & 0.104495 & 0.042343 \\
ALL & 0.456750 & 0.187555 & 0.026422 & 0.188537 & 0.104438 & 0.042370 \\
\bottomrule
\end{tabular}
\end{table}

\end{document}